\documentclass{article}

\usepackage{microtype}
\usepackage{graphicx}
\usepackage{multicol}
\usepackage{multirow}
\usepackage{makecell}
\usepackage{booktabs} 
\usepackage{xcolor}
\usepackage{subcaption}
\usepackage{enumitem}
\usepackage{colortbl}

\colorlet{darkgreen}{green!50!black}
\definecolor{royalblue}{RGB}{65,105,225}
\definecolor{reference}{RGB}{128,128,128}
\usepackage[breaklinks=true,colorlinks,citecolor=reference,bookmarks=false]{hyperref}

\usepackage[accepted]{icml2024}

\usepackage{amsmath}
\usepackage{amssymb}
\usepackage{mathtools}
\usepackage{amsthm}
\usepackage{pifont}

\usepackage{listings}
\usepackage{tcolorbox}
\newtcolorbox{dialogbox}{
  arc=4mm,
  colback=blue!3,
  colframe=black,
  rounded corners,
  boxrule=0.5pt,
  fonttitle=\bfseries,
  coltitle=black,
}
\usepackage{adjustbox}

\usepackage[capitalize,noabbrev]{cleveref}

\theoremstyle{plain}

\theoremstyle{definition}

\theoremstyle{remark}

\usepackage[textsize=tiny]{todonotes}

\def\ourmethod{EoH}

\begin{document}
\newcommand{\xl}[1]{{\color{orange}{[xl:#1]}}}
\newcommand{\fei}[1]{{\color{purple}{[fei:#1]}}}
\newcommand{\luz}[1]{{\color{magenta}{[zhichao:#1]}}}

\twocolumn[
\icmltitle{Evolution of Heuristics: Towards Efficient Automatic Algorithm Design Using Large Language Model}

\begin{icmlauthorlist}
\icmlauthor{Fei Liu}{cityu}
\icmlauthor{Xialiang Tong}{comp}
\icmlauthor{Mingxuan Yuan}{comp}
\icmlauthor{Xi Lin}{cityu}
\icmlauthor{Fu Luo}{sustech}
\icmlauthor{Zhenkun Wang}{sustech}
\icmlauthor{Zhichao Lu}{cityu}
\icmlauthor{Qingfu Zhang}{cityu}


\end{icmlauthorlist}

\icmlaffiliation{cityu}{Department of Computer Science, City University of Hong Kong}
\icmlaffiliation{comp}{Huawei Noah’s Ark Lab}
\icmlaffiliation{sustech}{School of System Design and
Intelligent Manufacturing, Southern University of Science and Technology}

\icmlcorrespondingauthor{Qingfu Zhang}{qingfu.zhang@cityu.edu.hk}
\icmlcorrespondingauthor{Zhenkun Wang}{wangzk3@sustech.edu.cn}



\vskip 0.3in
]



\printAffiliationsAndNotice{}  

\begin{abstract}

Heuristics are widely used for dealing with complex search and optimization problems. However, manual design of heuristics can be often very labour extensive and requires rich working experience and knowledge. This paper proposes Evolution of Heuristic (EoH), a novel evolutionary paradigm that leverages both Large Language Models (LLMs) and Evolutionary Computation (EC) methods for Automatic Heuristic Design (AHD). EoH represents the ideas of heuristics in natural language, termed \emph{thoughts}. They are then translated into executable \emph{codes} by LLMs. The evolution of both thoughts and codes in an evolutionary search framework makes it very effective and efficient for generating high-performance heuristics. Experiments on three widely studied combinatorial optimization benchmark problems demonstrate that EoH outperforms commonly used handcrafted heuristics and other recent AHD methods including FunSearch. Particularly, the heuristic produced by EoH with a low computational budget (in terms of the number of queries to LLMs) significantly outperforms widely-used human hand-crafted baseline algorithms for the online bin packing problem.

\end{abstract}
\section{Introduction}\label{sec:intro}

\begin{figure}[t]
    \begin{subfigure}[b]{0.23\textwidth}
    \centering
    \includegraphics[width=0.9\textwidth]{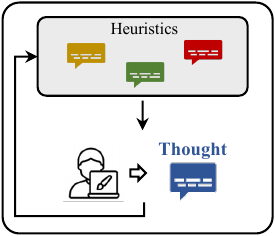}
    \caption{Manual Design}
    \end{subfigure} \hfill
    \begin{subfigure}[b]{0.23\textwidth}
    \centering
    \includegraphics[width=0.9\textwidth]{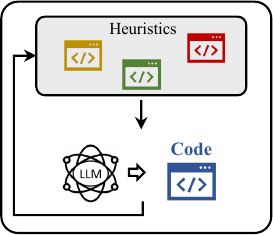}
    \caption{Evol. of codes (FunSearch)}
    \end{subfigure}\hfill
    \centering
    \begin{subfigure}[b]{0.48\textwidth}
    \centering
    \includegraphics[width=0.9\textwidth]{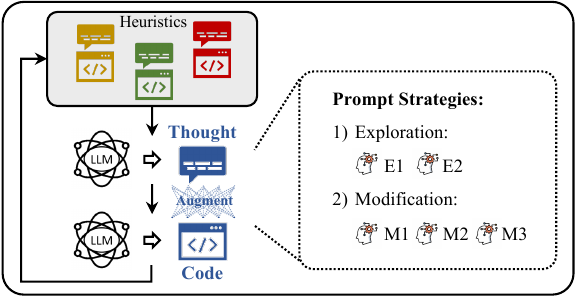}
    \caption{Evolution of both thoughts and codes (EoH, ours)}
    \end{subfigure}
    \caption{Heuristic design often (a) relies on human expertise with reasoning over thoughts; recent progress has been made on (b) search over the space of codes; while (c) our method evolves both \textit{thoughts} and \textit{codes} using large language models.\label{fig:enter-label}} 
\end{figure}

Heuristics are commonly used for tackling complex search and optimization problems. 
Over the last several decades, much effort has been devoted to designing effective heuristics, leading to simulated annealing~\cite{van1987simulated}, tabu search~\cite{glover1998tabu}, and iterated local search~\cite{lourencco2003iterated}, among many other methods~\cite{mart2018handbook}. 
These hand-crafted methods have been successfully used in a wide spectrum of real-world applications. However, different applications may require different algorithms and/or algorithm configurations. Manually designing, modifying, and configuring a heuristic for a given problem can be very labor-intensive and demands rich expert experience. This is a bottleneck in many application domains. To address this issue, Automatic Heuristic Design (AHD) has been proposed and become an active research area~\cite{burke2013hyper,stutzle2019automated}. 
%
AHD selects, tunes, or constructs effective heuristics for a given problem class automatically.  
Genetic Programming (GP) has been used in AHD~\cite{langdon2013foundations,zhang2023survey}. GP requires a set of permissible primitives or mutation operations for defining and generating heuristics. It could be very difficult to construct a suitable set in practice~\cite{o2010open}.


It is believed that Large Language Models (LLMs) ~\cite{chen2021evaluating, austin2021program, li2023starcoder} could be a powerful tool for generating new ideas and heuristics. 
%
%
However, standalone LLMs with prompt engineering can be insufficient for producing novel and useful ideas beyond existing knowledge~\cite{mahowald2023dissociating}.
Some attempts have been made to couple LLMs with Evolutionary Computation (EC) methods to produce heuristics in an automatic manner~\cite{yang2023large, meyerson2023language, chen2023evoprompting}. 
One representative work is FunSearch~\cite{romera2024mathematical}. 
It models AHD as a search problem in the space of functions, where each function is a heuristic represented by a program and it uses LLMs in an evolutionary framework to iteratively improve the quality of generated functions. FunSearch has been applied on several problems with great success. However, its mechanism is not very efficient and it needs a very large amount of computational resources to generate a quality heuristic.

In this paper, we present a new evolutionary paradigm, dubbed \emph{Evolution of Heuristic (\ourmethod{})}~\footnote{Our work and its preliminary version~\cite{liu2023algorithm} were developed independently of \citet{romera2024mathematical}.}, to take advantage of both LLMs and EC for AHD.
%
Specifically, we leverage a linguistic description, referred to as a \emph{thought}, to represent a high-level idea (i.e., key logic) of a heuristic. 
Then, a corresponding \emph{code} representation, i.e., an executable implementation of a heuristic, is generated via an LLM.
We propose an evolutionary framework to simultaneously evolve the thoughts and codes of heuristics in a cooperative manner.
We demonstrate that the LLM-assisted evolution of both thoughts and codes with curated prompts leads to state-of-the-art AHD performance.
We expect that \ourmethod{} serves as a step towards efficient and automatic algorithm design. 

In summary, our contributions are as follows:
\begin{itemize}
    \item We propose \ourmethod{}, a novel paradigm that uses LLMs to evolution both thoughts and codes for the automatic design of heuristics with minimum hand-craft design and no domain model training.
    %
    \item We develop several simple yet effective prompt strategies to guide LLMs toward generating more diverse and effective heuristics. These prompt strategies are generally applicable to other LLM-assisted search methods.  
    \item We comprehensively evaluate \ourmethod{} on three widely-studied combinatorial optimization benchmark problems.  We demonstrate that \ourmethod{} outperforms many existing AHD methods. In particular, \ourmethod{} identifies heuristics with better performance than those designed by FunSearch. \ourmethod{} uses much fewer queries to LLMs than FunSearch on online bin packing problem.

\end{itemize}
\vspace{1pt}

\section{Background and Related Works}\label{sec:related}

\subsection{Automatic Heuristic Design}
Automatic heuristic algorithm design is commonly known as hyper-heuristics~\cite{burke2013hyper,burke2019classification,stutzle2019automated}. With various effective methodologies~\cite{blot2016mo,lopez2016irace,akiba2019optuna} and frameworks~\cite{burke2019classification}, one can tune heuristics or combine different algorithmic components in an automatic manner. Much effort has been made to use machine learning techniques in automatic algorithm design~\cite{bengio2021machine,chen2022learning,he2021automl,li2023survey}. Among them, genetic programming~\cite{mei2022explainable,jia2022learning} provides an explainable approach to algorithm design. However, it requires hand-crafted algorithmic components and domain knowledge.

\subsection{LLMs for Heuristic Design}
Over the last few years, the ability of large language models has increased significantly~\cite{naveed2023comprehensive}. Recently, some effort has been made to use LLMs as basic algorithmic components to improve the performance of algorithms~\cite{yang2023large,guo2023towards}. Most of these works adopt LLM as optimizers~\cite{yang2023large} to directly generate new trial solutions through in-context learning. This approach faces challenges when applied to complex problems with large search space~\cite{yang2023large,nasir2023llmatic,zhao2023large,liu2023large}. Others integrate LLMs to assist algorithm design to extract deep algorithm features for heuristic selection~\cite{wu2023llm}, provide a guide for heuristic~\cite{shah2023navigation}, and design an algorithmic component~\cite{xiao2023llm}.  However, designing a competitive heuristic is still a challenge for standalone LLMs with prompt engineering.

\subsection{LLMs + EC}
Evolutionary computation is a generic optimization principle inspired by natural evolution~\cite{back1997handbook,eiben2015evolutionary}. Integration of EC in the prompt engineering of LLMs is very promising in improving performance in various domains~\cite{guo2023connecting,lehman2023evolution,wu2024evolutionary}. Evolutionary methods have been adopted in both code generation~\cite{liventsev2023fully,ma2023eureka,Lehman2024,hemberg2024evolving} and text generation~\cite{guo2023connecting,fernando2023promptbreeder,xu2023wizardlm}. The most related work to our effort is FunSearch~\cite{romera2024mathematical}, an evolutionary framework with LLMs to search functions automatically. Algorithms generated by FunSearch outperform hard-crafted algorithms on some optimization problems. However, FunSearch is computationally expensive and usually needs to generate millions of programs (i.e., queries to LLMs) to identify an effective heuristic function, which is not very practical for many users.

\vspace{1pt}
\section{Evolution of Heuristics (\ourmethod{})} \label{sec:method}

\subsection{Main Idea}

EoH aims at evolving both thoughts and codes to mimic the heuristic development conducted by human experts for efficient automatic heuristic design. To achieve this goal, EoH 
\begin{itemize}
    \item maintains both a natural language description and its corresponding code implementation for each heuristic. In each trial, it allows LLMs to first generate a heuristic in terms of natural language and then generate its corresponding code. The natural language description summarizes the main idea and provides a high-level understanding, while the code provides implementation details and settings that supplement the high-level thought.
    \item employs several prompt strategies to guide LLMs to do reasoning over existing thoughts and codes. These strategies are designed to learn from previous experiences and effectively explore the heuristic space. They can be regarded as fine-grained in-context learning approaches that combine thoughts and codes for heuristic search.
    \item evolves a population of candidate heuristics.  It uses LLMs in genetic operators such as crossover and mutation to produce new heuristics. Selection is also used to direct the search. The quality of each heuristic is evaluated on a set of problem instances.     
\end{itemize}
Unlike most evolutionary algorithms where individuals are candidate solutions to an optimization problem, An individual in \ourmethod{} is a heuristic designed for solving a given problem. We believe that the evolution of ``thoughts" should be an important research direction.      


 \ourmethod{} Integrates LLMs into an evolutionary framework. It generates and refines heuristics automatically. Unlike some classic automatic heuristic design methods~\cite{burke2013hyper}, \ourmethod{} doesn't need any 
 hand-crafted heuristic components or train new models.
 

\ourmethod{} evolves both thoughts and codes. Thoughts in natural language and designed prompt strategies enable \ourmethod{} to generate more diverse and effective heuristics. In contrast, FunSearch performs evolution of codes only and does not use prompt strategies explicitly.

\subsection{Evolution Framework}

\ourmethod{} maintains a population of $N$ heuristics, denoted as $P = \{h_1, \dots, h_N\}$, at each generation. Each heuristic $h_i$ is evaluated on a set of problem instances and assigned a fitness value $f(h_i)$.  

Five prompt strategies are designed to generate new heuristics. At each generation, each strategy is called $N$ times to generate $N$ heuristics. Each newly generated heuristic will be evaluated on problem instances and added to the current population if it is feasible. At most $5N$ new heuristics will be added to the current population at each generation. Then, $N$ best individual solutions from the current population will be selected to form the population for the next generation. 

EoH are summarised as follows: 

\textbf{Step 0 Initialization:} Initialize the population $P$ of $N$ heuristics $h_1, \ldots, h_N$ by prompting LLMs using \textbf{Initialization prompt}, its detail can be found in Section 3.4.

\textbf{Step 1 Generation of Heuristics}: If the stopping condition is not met, five \textbf{Evolution prompt strategies} (detailed in Section 3.4) are used simultaneously to generate $5N$ new heuristics. For each of the five prompt strategies, repeat the following process $N$ times: 
\begin{itemize}
    \item Step 1.1: Select parent heuristic(s) from the current population to construct a prompt for the strategy.
    \item Step 1.2: Request LLM  to generate a new heuristic as well as its corresponding code implementation. 
    \item Step 1.3: Evaluate the new heuristic on a set of evaluation instances to determine its fitness value. 
    \item Step 1.4: Add the new heuristic to the current population if the heuristic and code are feasible.
\end{itemize}

\textbf{Step 2 Population Management:} Select the $N$ best individual heuristics from the current population to form a population for the next generation. Go to \textbf{Step 1}.


\subsection{Heuristic Representation}
Each heuristic consists of three parts: 1) its description in natural language, 2) a code block in a pre-defined format, and 3) a fitness value.

The heuristic description comprises a few sentences in natural language. It is created by LLMs and presents a high-level thought.

The code block is an implementation of the heuristic. It should follow a pre-defined format so that it can be identified and seamlessly integrated into \ourmethod{} framework. In the experiments, we choose to implement it as a Python function. Three basic components should be explicitly given to format the code block: 1) Name of the function, 2) Input variables, and 3) Output variables.

The evaluation of heuristics in \ourmethod{} involves running the resulting algorithms on an instance set of the problem in question. This evaluation process differs from traditional evolutionary algorithms, which typically evaluate the objective function in a single instance. It is similar to some AHD approaches~\cite{lopez2016irace,hutter2011sequential} and is often costly.

\subsection{Prompt Strategies}

\paragraph{Initialization prompt}

In our experiments, we use LLMs to create all the initial heuristics, eliminating the need for expert knowledge. We inform the LLMs of the heuristic design task and instruct it to design a new heuristic by first presenting the description of the heuristic and then implementing it as a Python code block. The details of prompts for each problem are listed in the corresponding subsections in the \textbf{Appendix}. We repeat $N$ times to generate $N$ initial heuristics. 


\paragraph{Evolution prompts}
Five prompt strategies are proposed for creating new heuristics during evolution to mimic the heuristic development by humans. They are categorized into two groups: \textbf{E}xploration (\textbf{E1}, \textbf{E2}) and \textbf{M}odification (\textbf{M1}, \textbf{M2}, \textbf{M3}). The exploration strategies focus more on the exploration of the space of heuristics by conducting crossover-like operators on parent heuristics. 
The modification strategies refine a parent heuristic by modifying, adjusting parameters, and removing redundant parts. The details of these evolutionary prompts are listed as follows:

\textbf{E1:} Generate new heuristics that are as much different as possible from parent heuristics. First, $p$ heuristics are selected from the current population. Then, LLM is prompted to design a new heuristic that is different from these selected heuristics as much as possible in order to explore new ideas. 

\textbf{E2:} Explore new heuristics that share the same idea as the selected parent heuristics. First, $p$ heuristics are selected from the current population. Then, LLM is instructed to identify common ideas behind these heuristics. Then, a new heuristic is designed that are based the common ideas but are as much different as possible from the selected parents by introducing new parts. 

\textbf{M1:} Modify one heuristic for better performance. Firstly, one heuristic is selected from the population. Then, LLM is prompted to modify it to produce a new heuristic. 

\textbf{M2:} Modify the parameters of one selected heuristic. First, one heuristic is selected from the current population. Then, LLM is prompted to try different parameters in the current heuristic instead of designing a new one. 

\textbf{M3:} Simplify heuristics by removing redundant components. First, one heuristic is selected from the current population. Then, LLM is prompted to analyze and identify the main components in the selected heuristic and analyze whether there are any redundant components.
Finally, LLM is prompted to simplify the code implementation of the heuristic based on its analysis.

In all the above prompts, LLM is asked to first describe the heuristic and then provide a code implementation in a pre-defined format.

Any selection method can be used in EoH. In our experimental studies, all the heuristics in the current population are ranked according to their 
fitness.  Heuristic $i$ in the current population is randomly selected with probability $p_i \propto 1/(r_i+N)$, where $r_i$ is its rank and $N$ is the population size. 

\vspace{1pt}

\section{Experiments}\label{sec:experiment}

\subsection{Experimental Settings}
\paragraph{Benchmarks and datasets.} We consider three well-studied combinatorial optimization benchmark problems: 
\begin{itemize}
    \item \textit{Online bin packing problem}. The objective is to allocate a collection of items of different sizes into the fewest possible bins with a fixed capacity $C$. We focus on the online scenario~\cite{seiden2002online}, where items are packed as they arrive, in contrast to the offline scenario where all items are known beforehand. The instances that are used for evaluation during heuristic evolution are five Weibull instances of size 5k with a capacity of 100~\cite{romera2024mathematical}. The fitness value is set as the average $\frac{lb}{n}$ on the five instances, where $lb$ represents the lower bound of the optimal number of bins computed as in~\citet{martello1990lower} and $n$ is the number of bins used to pack all the items by the evaluated heuristic. 

    Methods in comparison include 
    \begin{itemize}
        \item human hand-crafted heuristics: the first fit and best fit heuristics~\cite{romera2024mathematical}. The first fit heuristic assigns the incoming item to the first bin that has sufficient available space, while the best fit heuristic selects the bin with the least available space that can still accommodate the item. 
        \item heuristics generated automatically: FunSearch~\cite{romera2024mathematical} is considered due to its excellent performance. We directly use the heuristic generated by FunSearch~\cite{romera2024mathematical} for comparison.
    \end{itemize}
    \item \textit{Traveling Salesman Problem (TSP)~\cite{matai2010traveling}}. It is to find the shortest route to visit all the given locations once and return to the starting location. It is one of the most widely-studied combinatorial optimization problems and a commonly used test bed for heuristics. The heuristic evolution process is conducted on a set of 64 TSP100 instances. The locations in these instances are randomly sampled from $[0,1]^2$~\cite{kool2018attention}. The average gap from the optimal solution (which is generated by Concorde~\cite{applegate2006concorde}) is used as the fitness value.  
    
    Methods in comparison include 
    \begin{itemize}
    \item hand-crafted heuristics: The nearest insertion\cite{rosenkrantz1977analysis} and farthest insertion~\cite{rosenkrantz1977analysis}, two commonly used constructive heuristics are used in comparison. Google Or-Tools~\cite{ortools}, one of the most popular solvers is also used.  We use the default settings of Or-Tools and the local search option suggested in OR-Tools to improve the solution quality. The stopping criterion is 60s for each instance.
    
    \item heuristics designed automatically by AI methods: they are the attention model (AM)~\cite{kool2018attention}, POMO~\cite{kwon2020pomo} and LEHD~\cite{luo2023neural}. AM~\cite{kool2018attention} is a seminal and well-known method for using neural networks to learn heuristics for combinatorial optimization. POMO~\cite{kwon2020pomo} adopts AM ideas and achieves state-of-the-art results. LEHD~\cite{luo2023neural} is a new revision of AM with a different heavy decoder structure and is trained using supervised learning. 
    \end{itemize}
    
    \item \textit{Flow Shop Scheduling Problem (FSSP)~\cite{emmons2012flow}}. It is to schedule $n$ jobs on $m$ machines, where each job contains $m$ operations that must be performed in a predetermined order on the respective machine. The objective is to minimize the total schedule length, known as the makespan. In the perturbation flow-shop scheduling problem, the processing order remains consistent throughout each step and no machine can execute multiple operations simultaneously. During heuristic evolution, we conduct evolution on 64 randomly generated instances. Each instance consists of 50 jobs and 2 to 20 machines. The processing times of the jobs are randomly generated from a uniform distribution ranging from 0 to 1~\cite{pan2021deep}. The average makespan serves as the fitness value. 
    
    Methods in comparison include
    \begin{itemize}
    \item hand-crafted heuristics: They are GUPTA~\cite{gupta1971functional},  CDS~\cite{campbell1970heuristic}
    NEH~\cite{nawaz1983heuristic} and NEHFF~\cite{fernandez2014insertion}.
    GUPTA~\cite{gupta1971functional} and  CDS~\cite{campbell1970heuristic}
    are two classic methods for flow-shop scheduling.  NEH~\cite{nawaz1983heuristic} and NEHFF~\cite{fernandez2014insertion} are widely recognized efficient heuristics for this problem. 
    \item heuristics designed automatically: PFSPNet and PFSPNet\_NEH~\cite{pan2021deep} are used. They are two recently proposed end-to-end deep learning solvers for flow-shop scheduling.
    \end{itemize}
    \end{itemize}

\begin{figure*}[tbp]
    \centering
    \includegraphics[width=\textwidth]{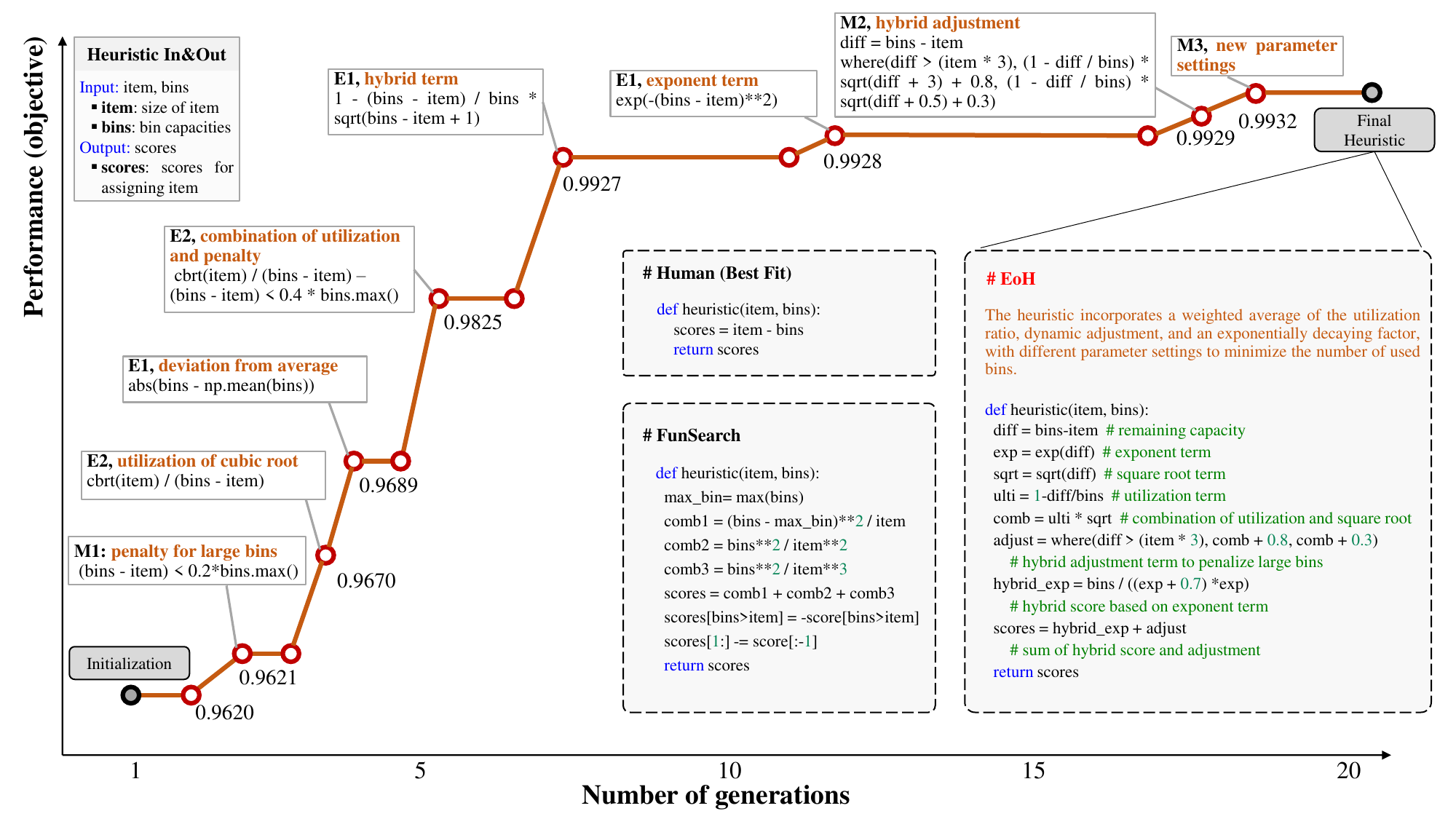}
    \caption{Evolution of \ourmethod{} for online bin packing. We outline the key \textbf{\textcolor{brown}{thoughts}} and the corresponding \textbf{code snippets} of the best heuristics produced in some generations during evolution. We list the prompt strategies. We present the best heuristic in the final population and compare it with the best fit heuristic and the heuristic produced by FunSearch.}
    \label{fig:convergence}
\end{figure*}

\paragraph{Implementation details.} 
\begin{itemize}
\item \textit{Online bin packing}. We adopt the settings in~\citet {romera2024mathematical} to design heuristics to determine the suitable bin allocation for incoming items~\cite{angelopoulos2023online}. Specifically, the task for \ourmethod{} is to design the scoring function for assigning items. The inputs are the size of the item and the rest capacities of bins. The output is the scores for the bins. The item will be assigned to the bin with the maximum score. 

\item \textit{Traveling salesman}. We use \ourmethod{} to design Guided Local Search (GLS) heuristics~\cite{alsheddy2018guided}. GLS is a strategy to help a local search to escape local optimal solutions. A key issue in GLS heuristics is how to update the objective function (i.e. landscape) to guide the local search to move to more promising areas.  Noting that the landscape is primarily determined by the distance matrix, the goal of \ourmethod{} is to produce a method for updating the distance matrix. 
Following~\cite{alsheddy2018guided,arnold2019knowledge}, the inputs are the distance matrix, the current route, and the number of edges. The output is an updated distance matrix. GLS runs local search operators on updated landscapes iteratively. Two local search operators used in our experiments are the \textit{relocate} and \textit{2-opt} operators. A detailed introduction of the GLS heuristics is presented in the \textbf{Appendix}.

\item  \textit{Flow shop scheduling}.
We use the \ourmethod{} to produce a heuristic for updating the objective landscape in GLS for this problem. 
The inputs in the heuristic are the time matrix, current scheduling, and the number of machines and jobs. The outputs are the updated time matrix and the calculated job perturbation priority. The local search operators used are the \textit{relocate} and \textit{swap} operators.
\end{itemize}

In our experiments, the number of generations in EoH for all problems is set to 20. The population size is 20 for online bin packing and 10 for TSP and FSSP. The number of parent heuristics used in E1 and E2 is $p=5$. The GPT-3.5-turbo pre-trained LLM is used. For TSP and FSSP, the maximum number of iterations for local search is set to 1,000, and the maximum running time for each instance is 60 seconds. The entire framework and the implementations of \ourmethod{} on the three problems are implemented in Python and executed on a single CPU i7-9700.

\subsection{Results}

\paragraph{Online bin packing.} 

We visualize the evolution of EoH for the online bin packing problem in Figure~\ref{fig:convergence}. We outline the key thoughts and the corresponding codes of the best heuristic found in some generations in the evolution process. We also list the prompt strategies that generate the thoughts and codes. The fitness value (objective) increases from 0.962 to 0.993 in 20 generations, which involves 2,000 LLM queries. 
As shown in Figure~\ref{fig:convergence}, in comparison to the best fit heuristic, the heuristics designed by both FunSearch and EoH are sophisticated.  The best heuristic found by EoH is a hybrid function with multiple components. It consists of remaining capacities, exponent term of remaining capacities, square root term of remaining capacities, and their combinations. It is the result of evolution. An example of the heuristic design using E2 prompt strategy is detailed in the \textbf{Appendix~\ref{e2}}. 

We test the best heuristic produced by EoH on instances of various sizes and capacities and compare it with two hand-crafted heuristics (i.e., first and best fit), and the best heuristic produced by FunSearch. The problem size ranges from 1k to 10k and the capacities are 100 and 500. Each set includes 5 randomly generated instances.
Table~\ref{tab:bin_pack} presents the average gaps to the lower bounds, where the best results are highlighted in bold. Our method is the best except for the results on the 10k\_C100 set (instances with a problem size of 10k and a capacity of 100). EoH makes only a few thousands of LLM queries, which is much faster than FunSearch (around 1 million queries reported in~\citet{romera2024mathematical}). Furthermore, our method achieves the same best gap in the training distribution and demonstrates excellent generalization performance to out-of-distribution instances. For instance, on the instance set with 1k items and a capacity of 500, FunSearch performs worse than two hand-crafted heuristics, whereas our method achieves the best gap of 2.13\%. More results are provided in the \textbf{Appendix~\ref{more_results_bp}}

\begin{table}[t]
\centering
\caption{\textbf{Online bin packing results.} Comparison of the fraction of excess bins to lower bound (lower is better) for various bin packing heuristics on Weibull instances. \label{tab:bin_pack}}
\renewcommand{\arraystretch}{1.3}
\resizebox{.48\textwidth}{!}{%
\begin{tabular}{@{\hspace{1mm}}lcccccc@{\hspace{1mm}}}
\toprule
 & 1k\_C100 & 5k\_C100 & 10k\_C100 & 1k\_C500 & 5k\_C500 & 10k\_C500 \\ \midrule
First Fit & 5.32\%          & 4.40\%          & 4.44\%          & 4.97\%          & 4.27\%          & 4.28\%          \\
Best Fit  & 4.87\%          & 4.08\%          & 4.09\%          & 4.50\%          & 3.91\%          & 3.95\%          \\
\midrule
FunSearch & 3.78\%          & \textbf{0.80\%} & \textbf{0.33\%} & 6.75\%          & 1.47\%          & 0.74\%          \\
 \midrule
\rowcolor{lightgray!40}
\ourmethod{} (ours)       & \textbf{2.24\%} & \textbf{0.80\%} & 0.61\%          & \textbf{2.13\%} & \textbf{0.78\%} & \textbf{0.61\%} \\ \bottomrule

\end{tabular}%
}
\end{table}

\begin{table}[t]
\centering
\small
\caption{\textbf{Traveling salesman problem results.} Comparison of the relative distance (\%) to the best-known solutions (lower is better) for various routing heuristics on a subset of TSPLib instances. \label{tab:tsp}}
\renewcommand{\arraystretch}{1.25}
\resizebox{.48\textwidth}{!}{%
\begin{tabular}{@{\hspace{1mm}}lcccccc@{\hspace{1mm}}}
\toprule
 & rd100 & pr124 & bier127 & kroA150 & u159  & kroB200 \\
 \midrule
NI       & 19.91 & 15.50 & 23.21   & 18.17   & 23.59 & 24.10   \\
FI       & 9.38  & 4.43  & 8.04    & 8.54    & 11.15 & 7.54    \\
Or-Tools & 0.01  & 0.55  & 0.66    & 0.02    & 1.75  & 2.57    \\
\midrule
AM       & 3.41  & 3.68  & 5.91    & 3.78    & 7.55  & 7.11    \\
POMO     & 0.01  & 0.60  & 13.72   & 0.70    & 0.95  & 1.58    \\
LEHD     & 0.01  & 1.11  & 4.76    & 1.40    & 1.13  & 0.64    \\
\midrule
\rowcolor{lightgray!40}
\ourmethod{}(Ours)      & \textbf{0.01}  & \textbf{0.00}  & \textbf{0.42}    & \textbf{0.00 }   & \textbf{0.00 } & \textbf{0.20}   \\
\bottomrule
\end{tabular}%
}
\end{table}

\begin{table}[ht]
\centering
\large
\renewcommand{\arraystretch}{1.35}
\caption{\textbf{Flow shop scheduling problem results.} Comparison of the average relative makespan (\%) to the baseline  (lower is better) on Taillard instances. \label{tab:fssp}}
\resizebox{.48\textwidth}{!}{%
\begin{tabular}{ccccccc}
\toprule
 & n20m10 & n20m20 & n50m10 & n50m20 & n100m10 & n100m20 \\
 \midrule
GUPTA        & 23.42  & 21.79  & 20.11  & 22.78  & 15.03   & 21.00   \\
CDS          & 12.87  & 10.35  & 12.72  & 15.03  & 9.36    & 13.55  \\
NEH          & 4.05   & 3.06   & 3.47   & 5.48   & 2.07    & 3.58  \\ 
NEHFF        & 4.15   & 2.72   & 3.62   & 5.10   & 1.88    & 3.73  \\
\midrule
PFSPNet      & 14.78  & 14.69  & 11.95  & 16.95  & 8.21    & 16.47  \\ 
PFSPNet\_NEH & 4.04   & 2.96   & 3.48   & 5.05   & 1.72    & 3.56  \\ 
\midrule
\rowcolor{lightgray!40}
\ourmethod{} (ours)   & \textbf{0.30}   & \textbf{0.10}   & \textbf{0.19}   & \textbf{0.60}   & \textbf{0.14}    & \textbf{0.41}   \\
\bottomrule

\end{tabular}%
}
\end{table}

\begin{figure}[ht]
    \centering
    \subfloat[TSP]{\includegraphics[width=0.98\linewidth]{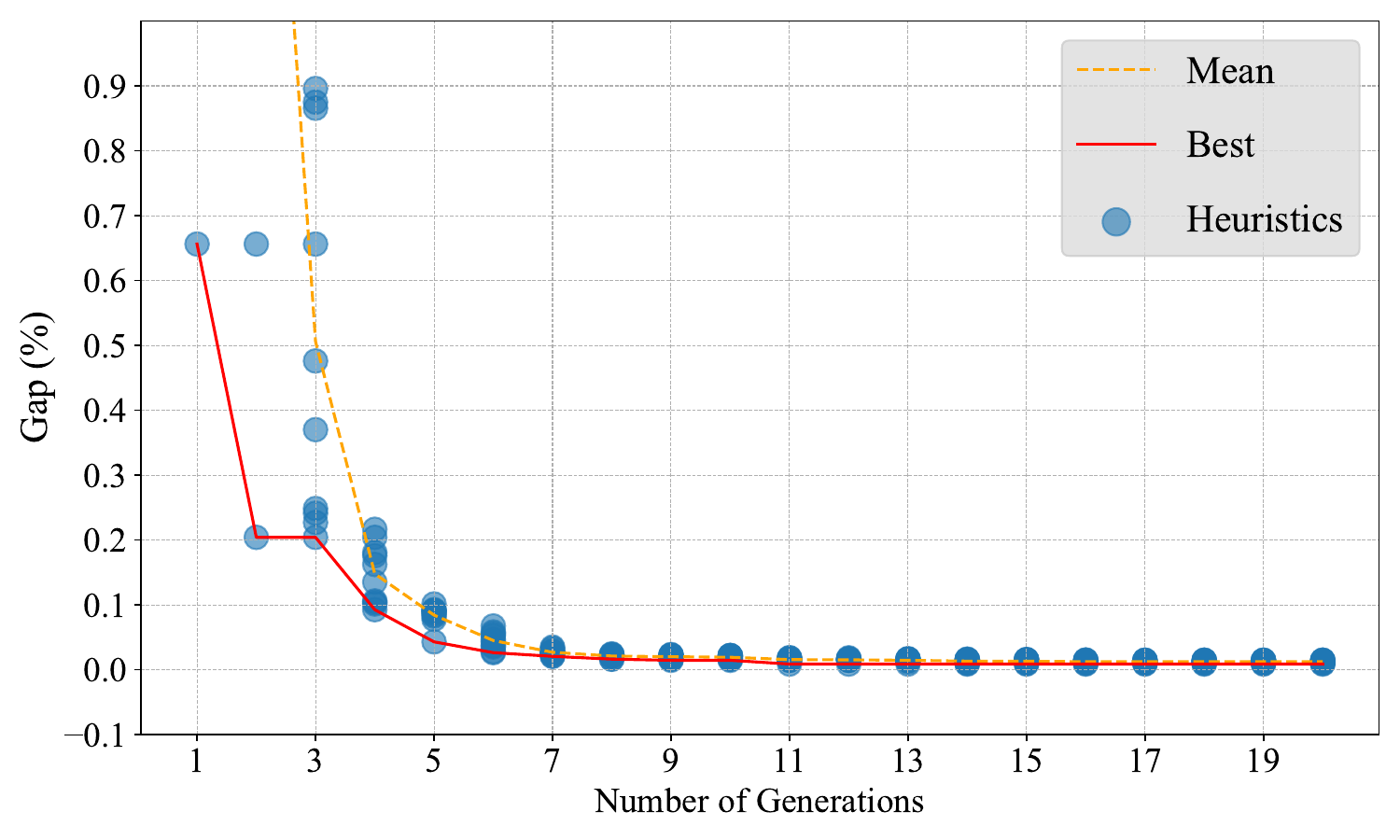}}
    \hfill
    \subfloat[FSSP]{\includegraphics[width=0.99\linewidth]{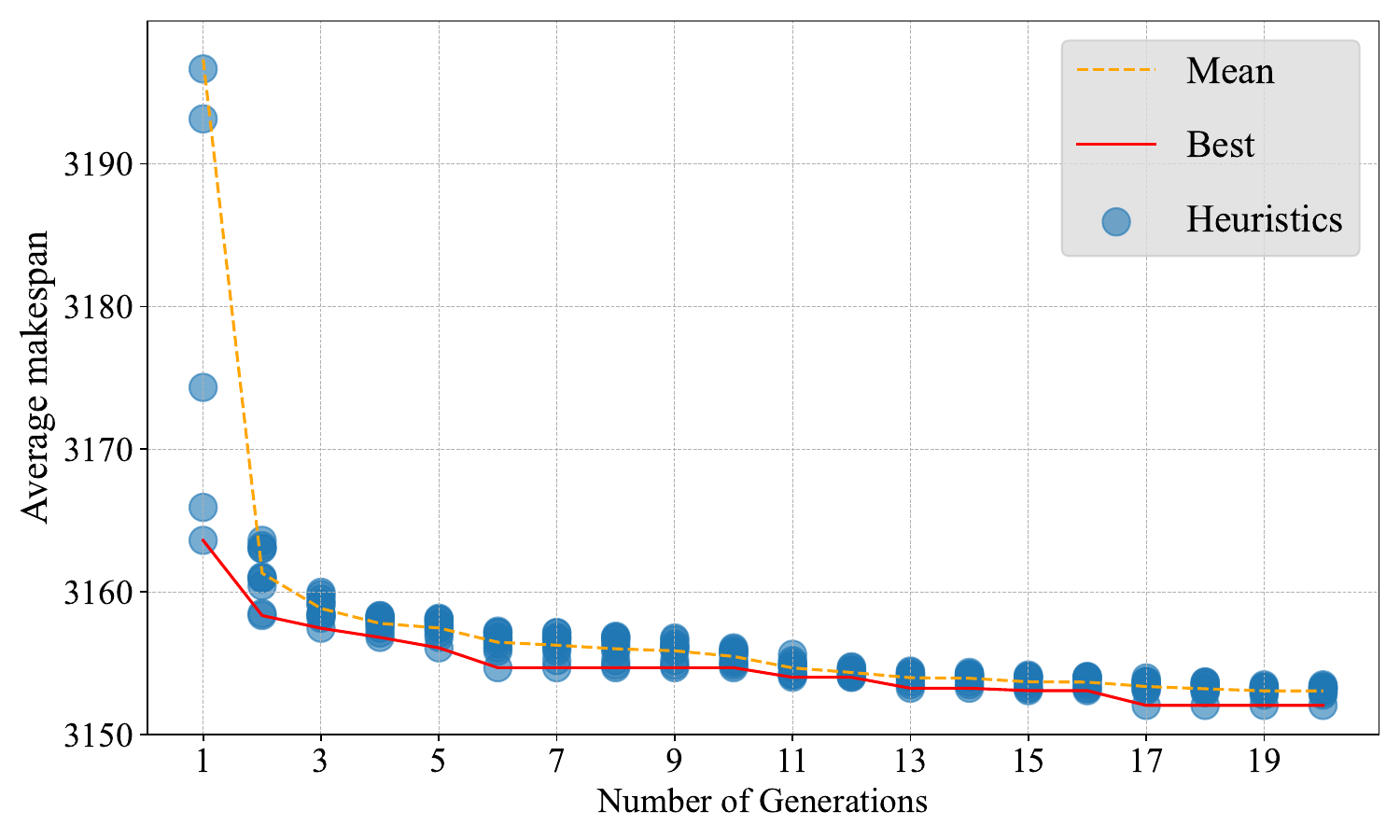}}
    \caption{Convergence curves of EoH on TSP and FSSP. The y-axis represents the average gap (\%) to the baseline and average makespan for TSP and FSSP, respectively. Each sample corresponds to a heuristic generated during the evolutionary process. The population size is 10 and the number of generations is 20. The gaps or makespans of some heuristics are out of the range of the figure or very close to each other. The red and orange lines depict the best and average performance of each population throughout the evolution.}
    \label{fig:evolution}
\end{figure}


\paragraph{Traveling salesman.}

Figure~\ref{fig:evolution} (a) illustrates the evolution process of heuristics in EoH for TSP. The y-axis represents the gap (\%) to the optimal solution and the x-axis represents the number of generations. Each blue data point represents a GLS heuristic produced by \ourmethod{}.   Every population consists of 10 heuristics. The red curve shows how the performance
of the best heuristic found at each generation is improved. 
The orange curve shows the average performance of the population at each generation. It is clear that EoH converges after about 20 generations.
We list the gap (\%) to the best-known solutions on six instances from TSPLib~\cite{reinelt1991tsplib} in Table~\ref{tab:tsp}. Comprehensive evaluation on other TSP instances and comparison with some other heuristics including hand-crafted GLS algorithms are provided in the \textbf{Appendix~\ref{more_results_tsp}}.

It should be pointed out that the heuristics produced by EoH consistently outperform other heuristics on all test instances. Notably, for pr124, kroA150, and u159, the EoH heuristic finds the best-known solutions (i.e., gap= 0\%). OR-Tools works well on average. However, its performance on large instances becomes worse perhaps due to the limited running time. The neural solvers are trained on the instances from a uniform distribution, which is the same as those used in EoH for fitness evaluation. These solvers can generate high-quality solutions on the instances from the same distribution. However, they deteriorate on out-of-distribution instances, such as the TSPLib instances. In contrast,  the heuristic designed by EoH works well on these out-of-distribution instances.


\paragraph{Flow shop scheduling.}
Figure~\ref{fig:evolution} (b) depicts the evolution process of heuristics on FSSP, demonstrating that it converges in about 20 generations. Table~\ref{tab:fssp} lists the experimental results of different heuristics on Taillard instances with the number of jobs ($n$) ranging from 20 to 100 and the number of machines ($m$) ranging from 10 to 20. The table displays the average gap (\%) to the upper bound provided in~\citet{taillard1993benchmarks}. Each set consists of 10 instances. The best results are highlighted in bold. \ourmethod{} produces the best results on all the test sets among all the heuristics. More results can be found in the \textbf{Appendix~\ref{more_results_fssp}}.

\subsection{Ablation Study}

An ablation study is carried out to provide a better understanding of the contribution of major components in EoH. We consider the following variants of EoH:
\begin{itemize}
    \item \ourmethod{}-e1: it doesn't use the three modification prompt strategies and E2, i.e. It uses only E1. 
    \item \ourmethod{}-e2: it doesn't use the three modification prompt strategies, i.e. It uses only E1 and E2. 
    \item EoC: it is the code-only variant of EoH. It doesn't include the thought part (i.e. the description of the heuristic in natural language). EoC only uses E1 (without linguistic description) to generate new codes. The number of parents $p=5$.
    \end{itemize}
    Table~\ref{tab:versions} summarizes the components used in various compared variants.
\begin{table}[t]
\centering
\caption{Comparison of different EoH variants on thoughts, codes, and prompt strategies in ablation study.}~\label{tab:versions}
\renewcommand{\arraystretch}{1.2}
\resizebox{0.4\textwidth}{!}{%
\begin{tabular}{lccc}
\toprule
\multicolumn{1}{l}{} & Thoughts  & Codes     & Prompt Strategies           \\
\midrule
EoC   & \ding{55} & \textcolor{red}{\ding{51}} & \textbf{E1}     \\
EoH-e1   & \textcolor{red}{\ding{51}} & \textcolor{red}{\ding{51}} & \textbf{E1}     \\
EoH-e2   & \textcolor{red}{\ding{51}} & \textcolor{red}{\ding{51}} & \textbf{E1, E2} \\
EoH      & \textcolor{red}{\ding{51}} & \textcolor{red}{\ding{51}} & \textbf{E1, E2, M1, M2, M3} \\
\bottomrule
\end{tabular}%
}
\end{table}

Experiments are carried out on online bin packing problem. The population size is 20. The GPT-3.5-turbo pre-trained LLM is used. The number of parent heuristics used in E1 and E2 is $p=5$. To have a fair comparison, all the variants use the same initial population. Consider that \ourmethod{} uses five prompt strategies while others use few strategies. We increase the numbers of generations correspondingly in each variant so that the total number of evaluated heuristics is the same.  
Table~\ref{tab:bin_pack_ablation} compares the final best heuristics produced by different variants. We can make the following observations: 
\begin{itemize}
    \item On average, EoH performs the best and EoH-e2 performs the second best. This implies that $M_1, M_2, M_3$, and $E_2$ do make positive contribution to the performance of EoH. 
    \item EoC performs the worst or the second worst on these test instances. It implies that the thought part in EoH is also very beneficial. 
\end{itemize}

\begin{table}[t]
\centering
\renewcommand{\arraystretch}{1.35}
\caption{Comparison of the gap to lower bound using heuristics designed by different versions on the six sets of Weibull instances. Each set consists of 5 instances sampled from the same distribution. In $a\_b$, $a$ means the number of items and $b$ means the capacity of bin.  
\label{tab:bin_pack_ablation}
}
\resizebox{.48\textwidth}{!}{%
\begin{tabular}{@{\hspace{1mm}}lcccccc@{\hspace{1mm}}}
\toprule
 & 1k\_100 & 5k\_100 & 10k\_100 & 1k\_500 & 5k\_500 & 10k\_500 \\ \midrule
EoC    & 148.63\%        & 3.23\%          & 24.55\%         & 150.89\%        & 12.53\%         & 32.02\%         \\
EoH-e1 & 4.13\%          & 0.99\%          & 0.60\%          & 58.17\%         & 55.48\%         & 54.79\%         \\
EoH-e2 & 4.28\%          & 0.97\%          & \textbf{0.56\%} & 5.86\%          & 1.36\%          & 0.73\%          \\
EoH    & \textbf{2.24\%} & \textbf{0.80\%} & 0.61\%          & \textbf{2.13\%} & \textbf{0.78\%} & \textbf{0.61\%} \\
\bottomrule

\end{tabular}%
}
\end{table}

\vspace{1pt}
\section{Discussion and Future Works}\label{sec:discussion}

\subsection{Discussion}
\paragraph{Interaction between Thoughts and Codes}
We carry out the following experiments to demonstrate the benefits of the use of both thoughts and codes in EoH. The thought and code can be regarded as multiple views of the heuristic. We will show that the multi-view thought and code representation contribute to EoH. 
We compare EoH with the following three variants: 
\begin{itemize}
    \item \textbf{C2C:} It uses only code to represent a heuristic. There is no natural language (thought) representation.
    \item \textbf{T2T2C:} It uses only thought for heuristic representation. Evolutionary prompts (i.e., E1-2, M1-3) only use thought representation to generate new heuristics. However, we still need to query LLMs to produce a code implementation for each heuristic for performance evaluation.  
    \item \textbf{T\&C2T2C:} It uses both thought and code representations of input heuristics in evolutionary prompts. We query LLMs with these prompts (i.e., E1-2, M1-3) to give only a natural language presentation of the produced heuristic. As in T2T2C, we need to request LLMs to produce a code implementation for each produced heuristic for performance evaluation.
    


\end{itemize}

Table~\ref{tab:interaction} lists comparison results (average gap (\%) to lb) of EoH with the above three variants on the online bin packing Weibull test instances. All experiments are carried out in the same experimental settings three times on the online bin packing problem.

It is evident that only use of codes (C2C) or thoughts (T2T2C) in evolutionary prompts is much worse than EoH. Thus, we can claim that the evolution of both codes and thoughts does make significant contribution to EoH. EoH is also better than T\&C2T2C. It implies that letting these prompts output both the codes and thoughts are helpful for the quality of the produced heuristics.  

\begin{table}[t]
\centering
\caption{Study of the effect of evolution of both thoughts and codes. Average gap (\%) to lb on online bin packing problem 5k Weibull instances. \label{tab:interaction}
}
\renewcommand{\arraystretch}{1.2}
\resizebox{0.4\textwidth}{!}{%
\begin{tabular}{lllll}
\toprule
 Setting          & Run 1 & Run 2 & Run 3 & Average \\
\midrule
 C2C      &  2.92	& 1.25	&3.53	&2.57 
   \\
 T2T2C    & 3.72  & 1.66  & 1.00  & 2.13       \\
T\&C2T2C    & 0.79  & 0.76  & 1.00  & 0.85      \\
 EoH                              & 0.68  & 0.67  & 0.62  & 0.66       \\
\bottomrule
\end{tabular}%
}
\end{table}

\paragraph{Different LLMs} 
We compare four commonly used LLMs: GPT3.5, Gemini Pro, CodeLlama, and Deepseek. All experiments are carried out in the same experimental settings three times on the online bin packing problem. In general, EoH can generate good-performance heuristics using these different LLMs. EoH using different LLMs with 2,000 queries to LLMs performs better than randomly querying GPT3.5 10,000 times. Nevertheless, our experimental results also show the benefits of using more powerful LLMs, e.g., GPT3.5 and Gemini Pro outperform other LLMs. 

\begin{table}[t]
\centering
\caption{Comparison of EoH with different LLMs. Average gap (\%) to lb on online bin packing problem 5k Weibull instances. \label{tab:LLM}}
\renewcommand{\arraystretch}{1.2}
\resizebox{0.48\textwidth}{!}{%
\begin{tabular}{llllll}
\toprule
Method & LLM           & Run 1 & Run 2 & Run 3 & Average \\
\midrule
Sampling      & GPT3.5  & 2.76  & 1.92  & 2.65  & 2.44  \\
EoH    & CodeLlama & 0.93  & 0.62  & 1.66  & 1.07        \\
EoH    & Deepseek      & 1.01  & 1.47  & 1.75  & 1.41       \\
EoH    & Gemini Pro    & 0.92  & 0.61  & 0.61  & 0.71         \\
EoH    & GPT3.5        & 0.68  & 0.67  & 0.62  & 0.66       \\
\bottomrule
\end{tabular}%
}
\end{table}
    
\paragraph{Use of Expert Heuristic}
We investigate the impact of the use of existing heuristics (expert heuristics) in EoH. Take the bin packing problem as an example, we adopt the heuristic provided in FunSearch paper~\cite{romera2024mathematical} as the existing expert heuristic and put it into the initial population of EoH. The rest of the initial heuristics are randomly generated. The results are given in Table~\ref{fig:expert}. We term the EoH with an expert in the initial population EoH expert and compare it to the original EoH and FunSearch. The results show that the adoption of elite expert heuristic benefits in the population benefits the final results in our test case. The EoH expert clearly surpasses both FunSearch and EoH. The knowledge of expert heuristics can be inherited and evolved during evolution to produce better heuristics. 

\begin{table}[t]
\centering
\caption{Study of the effect of using an existing expert heuristic in EoH. Average gap (\%) to lb on online bin packing problem 5k Weibull instances.}~\label{fig:expert}
\renewcommand{\arraystretch}{1.2}
\resizebox{0.4\textwidth}{!}{%
\begin{tabular}{lllll}
\toprule
Method     & Run 1 & Run 2 & Run 3 & Average \\
\midrule
FunSearch  & 0.94  & 0.82  & 1.15  & 0.97    \\
EoH        & 0.68  & 0.67  & 0.62  & 0.66    \\
EoH expert & 0.57  & 0.55  & 0.52  & 0.55   \\
\bottomrule
\end{tabular}
}
\end{table}

\subsection{Future Works}



It should be pointed out that the development of the evolution of heuristics using LLMs is still in its very early infancy. This paper and other research~\cite{romera2024mathematical} show that it is very promising for automatic algorithm design. Much effort should be made to advance this area. 

\paragraph{Pre-trained domain LLM}
Instead of using a general pre-trained LLM with linguistic and code generation capability, it is worthwhile studying how to train an LLM specifically for automatic algorithm design. Domain knowledge can be used for this purpose.  


\paragraph{Understanding of search space of heuristics}
EoH directly does its search on the space of heuristics. It is different from classic optimization algorithms which conduct their search in a well-defined math space such as $R^n$. It should be very important to study and understand search spaces of heuristics for further establishing theory and basic principles for the automatic design of algorithms.


\paragraph{Interaction with human experts}
A LLM in EoH can be regarded as an intelligent agent. During the process of EoH, It is straightforward to let human experts to replace LLM for generating, modifying, and evaluating heuristics at some stage. It should be interesting to study how to implement efficient and effective interaction with human experts in EoH. Ideas and techniques in collective intelligence~\cite{malone2022handbook} should be used for this purpose.


\vspace{1pt}
\section{Conclusion}\label{sec:conclusion}

This paper has proposed Evolution of Heuristics (EoH), which combines large language models (LLMs) and evolutionary computation (EC) methods to design heuristics in an automatic manner. By introducing the evolution of both thoughts and codes and using five prompt strategies, EoH mimics the process of heuristic design by human experts. We have tested EoH on three well-studied optimization problems, namely the online bin-packing problem, travel salesman problem, and flow shop scheduling problem. Experiments have shown that EoH can outperform human hand-crafted heuristics in some problem instances. EoH requires only a few thousand LLM requests while achieving better performance on most test instances. EoH offers a principled approach to automatic algorithm design. The source code can be found in \url{https://github.com/FeiLiu36/EoH}.

\newpage
\section*{Acknowledgements}
The work described in this paper was supported by the Research Grants Council of the Hong Kong Special Administrative Region, China (GRF Project No. CityU11215622), the National Natural Science Foundation of China (Grant No. 62106096), the Natural Science Foundation of Guangdong Province (Grant No. 2024A1515011759), the National Natural Science Foundation of Shenzhen (Grant No. JCYJ20220530113013031).

\section*{Impact Statement}

This paper presents work whose goal is to advance the field of Machine Learning. There are many potential societal consequences of our work, none of which we feel must be specifically highlighted here.

\vspace{1pt}




\bibliography{main}

\begin{thebibliography}{92}
\providecommand{\natexlab}[1]{#1}
\providecommand{\url}[1]{\texttt{#1}}
\expandafter\ifx\csname urlstyle\endcsname\relax
  \providecommand{\doi}[1]{doi: #1}\else
  \providecommand{\doi}{doi: \begingroup \urlstyle{rm}\Url}\fi

\bibitem[Akiba et~al.(2019)Akiba, Sano, Yanase, Ohta, and Koyama]{akiba2019optuna}
Akiba, T., Sano, S., Yanase, T., Ohta, T., and Koyama, M.
\newblock Optuna: A next-generation hyperparameter optimization framework.
\newblock In \emph{Proceedings of the 25th ACM SIGKDD international conference on knowledge discovery \& data mining}, pp.\  2623--2631, 2019.

\bibitem[Alsheddy et~al.(2018)Alsheddy, Voudouris, Tsang, and Alhindi]{alsheddy2018guided}
Alsheddy, A., Voudouris, C., Tsang, E.~P., and Alhindi, A.
\newblock Guided local search., 2018.

\bibitem[Angelopoulos et~al.(2023)Angelopoulos, Kamali, and Shadkami]{angelopoulos2023online}
Angelopoulos, S., Kamali, S., and Shadkami, K.
\newblock Online bin packing with predictions.
\newblock \emph{Journal of Artificial Intelligence Research}, 78:\penalty0 1111--1141, 2023.

\bibitem[Applegate et~al.(2006)Applegate, Bixby, Chvatal, and Cook]{applegate2006concorde}
Applegate, D., Bixby, R., Chvatal, V., and Cook, W.
\newblock Concorde tsp solver, 2006.

\bibitem[Arnold \& S{\"o}rensen(2019)Arnold and S{\"o}rensen]{arnold2019knowledge}
Arnold, F. and S{\"o}rensen, K.
\newblock Knowledge-guided local search for the vehicle routing problem.
\newblock \emph{Computers \& Operations Research}, 105:\penalty0 32--46, 2019.

\bibitem[Austin et~al.(2021)Austin, Odena, Nye, Bosma, Michalewski, Dohan, Jiang, Cai, Terry, Le, et~al.]{austin2021program}
Austin, J., Odena, A., Nye, M., Bosma, M., Michalewski, H., Dohan, D., Jiang, E., Cai, C., Terry, M., Le, Q., et~al.
\newblock Program synthesis with large language models.
\newblock \emph{arXiv preprint arXiv:2108.07732}, 2021.

\bibitem[B{\"a}ck et~al.(1997)B{\"a}ck, Fogel, and Michalewicz]{back1997handbook}
B{\"a}ck, T., Fogel, D.~B., and Michalewicz, Z.
\newblock Handbook of evolutionary computation.
\newblock \emph{Release}, 97\penalty0 (1):\penalty0 B1, 1997.

\bibitem[Bello et~al.(2016)Bello, Pham, Le, Norouzi, and Bengio]{bello2016neural}
Bello, I., Pham, H., Le, Q.~V., Norouzi, M., and Bengio, S.
\newblock Neural combinatorial optimization with reinforcement learning.
\newblock \emph{arXiv preprint arXiv:1611.09940}, 2016.

\bibitem[Bengio et~al.(2021)Bengio, Lodi, and Prouvost]{bengio2021machine}
Bengio, Y., Lodi, A., and Prouvost, A.
\newblock Machine learning for combinatorial optimization: a methodological tour d’horizon.
\newblock \emph{European Journal of Operational Research}, 290\penalty0 (2):\penalty0 405--421, 2021.

\bibitem[Blot et~al.(2016)Blot, Hoos, Jourdan, Kessaci-Marmion, and Trautmann]{blot2016mo}
Blot, A., Hoos, H.~H., Jourdan, L., Kessaci-Marmion, M.-{\'E}., and Trautmann, H.
\newblock Mo-paramils: A multi-objective automatic algorithm configuration framework.
\newblock In \emph{Learning and Intelligent Optimization: 10th International Conference, LION 10, Ischia, Italy, May 29--June 1, 2016, Revised Selected Papers 10}, pp.\  32--47. Springer, 2016.

\bibitem[Burke et~al.(2013)Burke, Gendreau, Hyde, Kendall, Ochoa, {\"O}zcan, and Qu]{burke2013hyper}
Burke, E.~K., Gendreau, M., Hyde, M., Kendall, G., Ochoa, G., {\"O}zcan, E., and Qu, R.
\newblock Hyper-heuristics: A survey of the state of the art.
\newblock \emph{Journal of the Operational Research Society}, 64:\penalty0 1695--1724, 2013.

\bibitem[Burke et~al.(2019)Burke, Hyde, Kendall, Ochoa, {\"O}zcan, and Woodward]{burke2019classification}
Burke, E.~K., Hyde, M.~R., Kendall, G., Ochoa, G., {\"O}zcan, E., and Woodward, J.~R.
\newblock A classification of hyper-heuristic approaches: revisited.
\newblock \emph{Handbook of metaheuristics}, pp.\  453--477, 2019.

\bibitem[Campbell et~al.(1970)Campbell, Dudek, and Smith]{campbell1970heuristic}
Campbell, H.~G., Dudek, R.~A., and Smith, M.~L.
\newblock A heuristic algorithm for the n job, m machine sequencing problem.
\newblock \emph{Management science}, 16\penalty0 (10):\penalty0 B--630, 1970.

\bibitem[Chen et~al.(2023)Chen, Dohan, and So]{chen2023evoprompting}
Chen, A., Dohan, D., and So, D.
\newblock Evoprompting: Language models for code-level neural architecture search.
\newblock In \emph{Advances in Neural Information Processing Systems}, 2023.

\bibitem[Chen et~al.(2021)Chen, Tworek, Jun, Yuan, Pinto, Kaplan, Edwards, Burda, Joseph, Brockman, et~al.]{chen2021evaluating}
Chen, M., Tworek, J., Jun, H., Yuan, Q., Pinto, H. P. d.~O., Kaplan, J., Edwards, H., Burda, Y., Joseph, N., Brockman, G., et~al.
\newblock Evaluating large language models trained on code.
\newblock \emph{arXiv preprint arXiv:2107.03374}, 2021.

\bibitem[Chen et~al.(2022)Chen, Chen, Chen, Wang, Heaton, Liu, and Yin]{chen2022learning}
Chen, T., Chen, X., Chen, W., Wang, Z., Heaton, H., Liu, J., and Yin, W.
\newblock Learning to optimize: A primer and a benchmark.
\newblock \emph{The Journal of Machine Learning Research}, 23\penalty0 (1):\penalty0 8562--8620, 2022.

\bibitem[Chu et~al.(2023)Chu, Chen, Chen, Yu, He, Wang, Peng, Liu, Qin, and Liu]{chu2023survey}
Chu, Z., Chen, J., Chen, Q., Yu, W., He, T., Wang, H., Peng, W., Liu, M., Qin, B., and Liu, T.
\newblock A survey of chain of thought reasoning: Advances, frontiers and future.
\newblock \emph{arXiv preprint arXiv:2309.15402}, 2023.

\bibitem[Deudon et~al.(2018)Deudon, Cournut, Lacoste, Adulyasak, and Rousseau]{deudon2018learning}
Deudon, M., Cournut, P., Lacoste, A., Adulyasak, Y., and Rousseau, L.-M.
\newblock Learning heuristics for the tsp by policy gradient.
\newblock In \emph{International conference on the integration of constraint programming, artificial intelligence, and operations research}, pp.\  170--181. Springer, 2018.

\bibitem[Drakulic et~al.(2023)Drakulic, Michel, Mai, Sors, and Andreoli]{drakulic2023bq}
Drakulic, D., Michel, S., Mai, F., Sors, A., and Andreoli, J.-M.
\newblock Bq-nco: Bisimulation quotienting for generalizable neural combinatorial optimization.
\newblock \emph{arXiv preprint arXiv:2301.03313}, 2023.

\bibitem[Eiben \& Smith(2015)Eiben and Smith]{eiben2015evolutionary}
Eiben, A.~E. and Smith, J.
\newblock From evolutionary computation to the evolution of things.
\newblock \emph{Nature}, 521\penalty0 (7553):\penalty0 476--482, 2015.

\bibitem[Emmons \& Vairaktarakis(2012)Emmons and Vairaktarakis]{emmons2012flow}
Emmons, H. and Vairaktarakis, G.
\newblock \emph{Flow shop scheduling: theoretical results, algorithms, and applications}, volume 182.
\newblock Springer Science \& Business Media, 2012.

\bibitem[Fernandez-Viagas \& Framinan(2014)Fernandez-Viagas and Framinan]{fernandez2014insertion}
Fernandez-Viagas, V. and Framinan, J.~M.
\newblock On insertion tie-breaking rules in heuristics for the permutation flowshop scheduling problem.
\newblock \emph{Computers \& Operations Research}, 45:\penalty0 60--67, 2014.

\bibitem[Fernando et~al.(2023)Fernando, Banarse, Michalewski, Osindero, and Rockt{\"a}schel]{fernando2023promptbreeder}
Fernando, C., Banarse, D., Michalewski, H., Osindero, S., and Rockt{\"a}schel, T.
\newblock Promptbreeder: Self-referential self-improvement via prompt evolution.
\newblock \emph{arXiv preprint arXiv:2309.16797}, 2023.

\bibitem[Fu et~al.(2021)Fu, Qiu, and Zha]{fu2021generalize}
Fu, Z.-H., Qiu, K.-B., and Zha, H.
\newblock Generalize a small pre-trained model to arbitrarily large tsp instances.
\newblock In \emph{Proceedings of the AAAI Conference on Artificial Intelligence}, volume~35, pp.\  7474--7482, 2021.

\bibitem[Glover \& Laguna(1998)Glover and Laguna]{glover1998tabu}
Glover, F. and Laguna, M.
\newblock \emph{Tabu search}.
\newblock Springer, 1998.

\bibitem[Guo et~al.(2023{\natexlab{a}})Guo, Chen, Tsai, and Lin]{guo2023towards}
Guo, P.-F., Chen, Y.-H., Tsai, Y.-D., and Lin, S.-D.
\newblock Towards optimizing with large language models.
\newblock \emph{arXiv preprint arXiv:2310.05204}, 2023{\natexlab{a}}.

\bibitem[Guo et~al.(2023{\natexlab{b}})Guo, Wang, Guo, Li, Song, Tan, Liu, Bian, and Yang]{guo2023connecting}
Guo, Q., Wang, R., Guo, J., Li, B., Song, K., Tan, X., Liu, G., Bian, J., and Yang, Y.
\newblock Connecting large language models with evolutionary algorithms yields powerful prompt optimizers.
\newblock \emph{arXiv preprint arXiv:2309.08532}, 2023{\natexlab{b}}.

\bibitem[Gupta(1971)]{gupta1971functional}
Gupta, J.~N.
\newblock A functional heuristic algorithm for the flowshop scheduling problem.
\newblock \emph{Journal of the Operational Research Society}, 22:\penalty0 39--47, 1971.

\bibitem[He et~al.(2021)He, Zhao, and Chu]{he2021automl}
He, X., Zhao, K., and Chu, X.
\newblock Automl: A survey of the state-of-the-art.
\newblock \emph{Knowledge-Based Systems}, 212:\penalty0 106622, 2021.

\bibitem[Helsgaun(2017)]{LKH3}
Helsgaun, K.
\newblock An extension of the lin-kernighan-helsgaun tsp solver for constrained traveling salesman and vehicle routing problems.
\newblock \emph{Roskilde: Roskilde University}, 12, 2017.

\bibitem[Hemberg et~al.(2024)Hemberg, Moskal, and O'Reilly]{hemberg2024evolving}
Hemberg, E., Moskal, S., and O'Reilly, U.-M.
\newblock Evolving code with a large language model.
\newblock \emph{arXiv preprint arXiv:2401.07102}, 2024.

\bibitem[Hudson et~al.(2021)Hudson, Li, Malencia, and Prorok]{hudson2021graph}
Hudson, B., Li, Q., Malencia, M., and Prorok, A.
\newblock Graph neural network guided local search for the traveling salesperson problem.
\newblock \emph{arXiv preprint arXiv:2110.05291}, 2021.

\bibitem[Hutter et~al.(2011)Hutter, Hoos, and Leyton-Brown]{hutter2011sequential}
Hutter, F., Hoos, H.~H., and Leyton-Brown, K.
\newblock Sequential model-based optimization for general algorithm configuration.
\newblock In \emph{Learning and Intelligent Optimization: 5th International Conference, LION 5, Rome, Italy, January 17-21, 2011. Selected Papers 5}, pp.\  507--523. Springer, 2011.

\bibitem[Jia et~al.(2022)Jia, Mei, and Zhang]{jia2022learning}
Jia, Y.-H., Mei, Y., and Zhang, M.
\newblock Learning heuristics with different representations for stochastic routing.
\newblock \emph{IEEE Transactions on Cybernetics}, 2022.

\bibitem[Joshi et~al.(2019)Joshi, Laurent, and Bresson]{joshi2019efficient}
Joshi, C.~K., Laurent, T., and Bresson, X.
\newblock An efficient graph convolutional network technique for the travelling salesman problem.
\newblock \emph{arXiv preprint arXiv:1906.01227}, 2019.

\bibitem[Kojima et~al.(2022)Kojima, Gu, Reid, Matsuo, and Iwasawa]{kojima2022large}
Kojima, T., Gu, S.~S., Reid, M., Matsuo, Y., and Iwasawa, Y.
\newblock Large language models are zero-shot reasoners.
\newblock \emph{Advances in neural information processing systems}, 35:\penalty0 22199--22213, 2022.

\bibitem[Kool et~al.(2018)Kool, Van~Hoof, and Welling]{kool2018attention}
Kool, W., Van~Hoof, H., and Welling, M.
\newblock Attention, learn to solve routing problems!
\newblock \emph{arXiv preprint arXiv:1803.08475}, 2018.

\bibitem[Kool et~al.(2022)Kool, van Hoof, Gromicho, and Welling]{kool2022deep}
Kool, W., van Hoof, H., Gromicho, J., and Welling, M.
\newblock Deep policy dynamic programming for vehicle routing problems.
\newblock In \emph{Integration of Constraint Programming, Artificial Intelligence, and Operations Research: 19th International Conference, CPAIOR 2022, Los Angeles, CA, USA, June 20-23, 2022, Proceedings}, pp.\  190--213. Springer, 2022.

\bibitem[Kwon et~al.(2020)Kwon, Choo, Kim, Yoon, Gwon, and Min]{kwon2020pomo}
Kwon, Y.-D., Choo, J., Kim, B., Yoon, I., Gwon, Y., and Min, S.
\newblock Pomo: Policy optimization with multiple optima for reinforcement learning.
\newblock \emph{Advances in Neural Information Processing Systems}, 33:\penalty0 21188--21198, 2020.

\bibitem[Kwon et~al.(2021)Kwon, Choo, Yoon, Park, Park, and Gwon]{kwon2021matrix}
Kwon, Y.-D., Choo, J., Yoon, I., Park, M., Park, D., and Gwon, Y.
\newblock Matrix encoding networks for neural combinatorial optimization.
\newblock In \emph{Advances in Neural Information Processing Systems}, 2021.

\bibitem[Langdon \& Poli(2013)Langdon and Poli]{langdon2013foundations}
Langdon, W.~B. and Poli, R.
\newblock \emph{Foundations of genetic programming}.
\newblock Springer Science \& Business Media, 2013.

\bibitem[Lehman et~al.(2023)Lehman, Gordon, Jain, Ndousse, Yeh, and Stanley]{lehman2023evolution}
Lehman, J., Gordon, J., Jain, S., Ndousse, K., Yeh, C., and Stanley, K.~O.
\newblock Evolution through large models.
\newblock In \emph{Handbook of Evolutionary Machine Learning}, pp.\  331--366. Springer, 2023.

\bibitem[Lehman et~al.(2024)Lehman, Gordon, Jain, Ndousse, Yeh, and Stanley]{Lehman2024}
Lehman, J., Gordon, J., Jain, S., Ndousse, K., Yeh, C., and Stanley, K.~O.
\newblock \emph{Evolution Through Large Models}, pp.\  331--366.
\newblock Springer Nature Singapore, Singapore, 2024.

\bibitem[Li et~al.(2023{\natexlab{a}})Li, Ma, Yu, Xue, Zhang, and Jin]{li2023survey}
Li, N., Ma, L., Yu, G., Xue, B., Zhang, M., and Jin, Y.
\newblock Survey on evolutionary deep learning: Principles, algorithms, applications, and open issues.
\newblock \emph{ACM Computing Surveys}, 56\penalty0 (2):\penalty0 1--34, 2023{\natexlab{a}}.

\bibitem[Li et~al.(2023{\natexlab{b}})Li, Allal, Zi, Muennighoff, Kocetkov, Mou, Marone, Akiki, Li, Chim, et~al.]{li2023starcoder}
Li, R., Allal, L.~B., Zi, Y., Muennighoff, N., Kocetkov, D., Mou, C., Marone, M., Akiki, C., Li, J., Chim, J., et~al.
\newblock Starcoder: may the source be with you!
\newblock \emph{arXiv preprint arXiv:2305.06161}, 2023{\natexlab{b}}.

\bibitem[Liu et~al.(2023{\natexlab{a}})Liu, Lin, Wang, Yao, Tong, Yuan, and Zhang]{liu2023large}
Liu, F., Lin, X., Wang, Z., Yao, S., Tong, X., Yuan, M., and Zhang, Q.
\newblock Large language model for multi-objective evolutionary optimization.
\newblock \emph{arXiv preprint arXiv:2310.12541}, 2023{\natexlab{a}}.

\bibitem[Liu et~al.(2023{\natexlab{b}})Liu, Tong, Yuan, and Zhang]{liu2023algorithm}
Liu, F., Tong, X., Yuan, M., and Zhang, Q.
\newblock Algorithm evolution using large language model.
\newblock \emph{arXiv preprint arXiv:2311.15249}, 2023{\natexlab{b}}.

\bibitem[Liventsev et~al.(2023)Liventsev, Grishina, H{\"a}rm{\"a}, and Moonen]{liventsev2023fully}
Liventsev, V., Grishina, A., H{\"a}rm{\"a}, A., and Moonen, L.
\newblock Fully autonomous programming with large language models.
\newblock \emph{arXiv preprint arXiv:2304.10423}, 2023.

\bibitem[Long(2023)]{long2023large}
Long, J.
\newblock Large language model guided tree-of-thought.
\newblock \emph{arXiv preprint arXiv:2305.08291}, 2023.

\bibitem[L{\'o}pez-Ib{\'a}{\~n}ez et~al.(2016)L{\'o}pez-Ib{\'a}{\~n}ez, Dubois-Lacoste, C{\'a}ceres, Birattari, and St{\"u}tzle]{lopez2016irace}
L{\'o}pez-Ib{\'a}{\~n}ez, M., Dubois-Lacoste, J., C{\'a}ceres, L.~P., Birattari, M., and St{\"u}tzle, T.
\newblock The irace package: Iterated racing for automatic algorithm configuration.
\newblock \emph{Operations Research Perspectives}, 3:\penalty0 43--58, 2016.

\bibitem[Louren{\c{c}}o et~al.(2003)Louren{\c{c}}o, Martin, and St{\"u}tzle]{lourencco2003iterated}
Louren{\c{c}}o, H.~R., Martin, O.~C., and St{\"u}tzle, T.
\newblock Iterated local search.
\newblock In \emph{Handbook of metaheuristics}, pp.\  320--353. Springer, 2003.

\bibitem[Luo et~al.(2023)Luo, Lin, Liu, Zhang, and Wang]{luo2023neural}
Luo, F., Lin, X., Liu, F., Zhang, Q., and Wang, Z.
\newblock Neural combinatorial optimization with heavy decoder: Toward large scale generalization.
\newblock \emph{arXiv preprint arXiv:2310.07985}, 2023.

\bibitem[Ma et~al.(2023)Ma, Liang, Wang, Huang, Bastani, Jayaraman, Zhu, Fan, and Anandkumar]{ma2023eureka}
Ma, Y.~J., Liang, W., Wang, G., Huang, D.-A., Bastani, O., Jayaraman, D., Zhu, Y., Fan, L., and Anandkumar, A.
\newblock Eureka: Human-level reward design via coding large language models.
\newblock \emph{arXiv preprint arXiv:2310.12931}, 2023.

\bibitem[Mahowald et~al.(2023)Mahowald, Ivanova, Blank, Kanwisher, Tenenbaum, and Fedorenko]{mahowald2023dissociating}
Mahowald, K., Ivanova, A.~A., Blank, I.~A., Kanwisher, N., Tenenbaum, J.~B., and Fedorenko, E.
\newblock Dissociating language and thought in large language models: a cognitive perspective.
\newblock \emph{arXiv preprint arXiv:2301.06627}, 2023.

\bibitem[Malone \& Bernstein(2022)Malone and Bernstein]{malone2022handbook}
Malone, T.~W. and Bernstein, M.~S.
\newblock \emph{Handbook of collective intelligence}.
\newblock MIT press, 2022.

\bibitem[Mart et~al.(2018)Mart, Pardalos, and Resende]{mart2018handbook}
Mart, R., Pardalos, P.~M., and Resende, M.~G.
\newblock \emph{Handbook of heuristics}.
\newblock Springer Publishing Company, Incorporated, 2018.

\bibitem[Martello \& Toth(1990)Martello and Toth]{martello1990lower}
Martello, S. and Toth, P.
\newblock Lower bounds and reduction procedures for the bin packing problem.
\newblock \emph{Discrete applied mathematics}, 28\penalty0 (1):\penalty0 59--70, 1990.

\bibitem[Matai et~al.(2010)Matai, Singh, and Mittal]{matai2010traveling}
Matai, R., Singh, S.~P., and Mittal, M.~L.
\newblock Traveling salesman problem: an overview of applications, formulations, and solution approaches.
\newblock \emph{Traveling salesman problem, theory and applications}, 1\penalty0 (1):\penalty0 1--25, 2010.

\bibitem[Mei et~al.(2022)Mei, Chen, Lensen, Xue, and Zhang]{mei2022explainable}
Mei, Y., Chen, Q., Lensen, A., Xue, B., and Zhang, M.
\newblock Explainable artificial intelligence by genetic programming: A survey.
\newblock \emph{IEEE Transactions on Evolutionary Computation}, 2022.

\bibitem[Meyerson et~al.(2023)Meyerson, Nelson, Bradley, Moradi, Hoover, and Lehman]{meyerson2023language}
Meyerson, E., Nelson, M.~J., Bradley, H., Moradi, A., Hoover, A.~K., and Lehman, J.
\newblock Language model crossover: Variation through few-shot prompting.
\newblock \emph{arXiv preprint arXiv:2302.12170}, 2023.

\bibitem[Nasir et~al.(2023)Nasir, Earle, Togelius, James, and Cleghorn]{nasir2023llmatic}
Nasir, M.~U., Earle, S., Togelius, J., James, S., and Cleghorn, C.
\newblock Llmatic: Neural architecture search via large language models and quality-diversity optimization.
\newblock \emph{arXiv preprint arXiv:2306.01102}, 2023.

\bibitem[Naveed et~al.(2023)Naveed, Khan, Qiu, Saqib, Anwar, Usman, Barnes, and Mian]{naveed2023comprehensive}
Naveed, H., Khan, A.~U., Qiu, S., Saqib, M., Anwar, S., Usman, M., Barnes, N., and Mian, A.
\newblock A comprehensive overview of large language models.
\newblock \emph{arXiv preprint arXiv:2307.06435}, 2023.

\bibitem[Nawaz et~al.(1983)Nawaz, Enscore~Jr, and Ham]{nawaz1983heuristic}
Nawaz, M., Enscore~Jr, E.~E., and Ham, I.
\newblock A heuristic algorithm for the m-machine, n-job flow-shop sequencing problem.
\newblock \emph{Omega}, 11\penalty0 (1):\penalty0 91--95, 1983.

\bibitem[O’Neill et~al.(2010)O’Neill, Vanneschi, Gustafson, and Banzhaf]{o2010open}
O’Neill, M., Vanneschi, L., Gustafson, S., and Banzhaf, W.
\newblock Open issues in genetic programming.
\newblock \emph{Genetic Programming and Evolvable Machines}, 11:\penalty0 339--363, 2010.

\bibitem[Pan et~al.(2021)Pan, Wang, Wang, and Lu]{pan2021deep}
Pan, Z., Wang, L., Wang, J., and Lu, J.
\newblock Deep reinforcement learning based optimization algorithm for permutation flow-shop scheduling.
\newblock \emph{IEEE Transactions on Emerging Topics in Computational Intelligence}, 2021.

\bibitem[Perron \& Furnon()Perron and Furnon]{ortools}
Perron, L. and Furnon, V.
\newblock Or-tools.
\newblock URL \url{https://developers.google.com/optimization/}.

\bibitem[Qiu et~al.(2022)Qiu, Sun, and Yang]{qiu2022DIMES}
Qiu, R., Sun, Z., and Yang, Y.
\newblock Dimes: A differentiable meta solver for combinatorial optimization problems.
\newblock \emph{arXiv preprint arXiv:2210.04123}, 2022.

\bibitem[Reinelt(1991)]{reinelt1991tsplib}
Reinelt, G.
\newblock Tsplib—a traveling salesman problem library.
\newblock \emph{ORSA journal on computing}, 3\penalty0 (4):\penalty0 376--384, 1991.

\bibitem[Romera-Paredes et~al.(2024)Romera-Paredes, Barekatain, Novikov, Balog, Kumar, Dupont, Ruiz, Ellenberg, Wang, Fawzi, et~al.]{romera2024mathematical}
Romera-Paredes, B., Barekatain, M., Novikov, A., Balog, M., Kumar, M.~P., Dupont, E., Ruiz, F.~J., Ellenberg, J.~S., Wang, P., Fawzi, O., et~al.
\newblock Mathematical discoveries from program search with large language models.
\newblock \emph{Nature}, 625\penalty0 (7995):\penalty0 468--475, 2024.

\bibitem[Rosenkrantz et~al.(1977)Rosenkrantz, Stearns, and Lewis]{rosenkrantz1977analysis}
Rosenkrantz, D.~J., Stearns, R.~E., and Lewis, II, P.~M.
\newblock An analysis of several heuristics for the traveling salesman problem.
\newblock \emph{SIAM journal on computing}, 6\penalty0 (3):\penalty0 563--581, 1977.

\bibitem[Seiden(2002)]{seiden2002online}
Seiden, S.~S.
\newblock On the online bin packing problem.
\newblock \emph{Journal of the ACM (JACM)}, 49\penalty0 (5):\penalty0 640--671, 2002.

\bibitem[Sel et~al.(2023)Sel, Al-Tawaha, Khattar, Wang, Jia, and Jin]{sel2023algorithm}
Sel, B., Al-Tawaha, A., Khattar, V., Wang, L., Jia, R., and Jin, M.
\newblock Algorithm of thoughts: Enhancing exploration of ideas in large language models.
\newblock \emph{arXiv preprint arXiv:2308.10379}, 2023.

\bibitem[Shah et~al.(2023)Shah, Equi, Osi{\'n}ski, Xia, Ichter, and Levine]{shah2023navigation}
Shah, D., Equi, M.~R., Osi{\'n}ski, B., Xia, F., Ichter, B., and Levine, S.
\newblock Navigation with large language models: Semantic guesswork as a heuristic for planning.
\newblock In \emph{Conference on Robot Learning}, pp.\  2683--2699. PMLR, 2023.

\bibitem[Shi et~al.(2018)Shi, Zhang, and Tsang]{shi2018eb}
Shi, J., Zhang, Q., and Tsang, E.
\newblock Eb-gls: an improved guided local search based on the big valley structure.
\newblock \emph{Memetic computing}, 10:\penalty0 333--350, 2018.

\bibitem[St{\"u}tzle(1998)]{stutzle1998applying}
St{\"u}tzle, T.
\newblock Applying iterated local search to the permutation flow shop problem.
\newblock Technical report, Technical Report AIDA-98-04, FG Intellektik, TU Darmstadt, 1998.

\bibitem[St{\"u}tzle \& L{\'o}pez-Ib{\'a}{\~n}ez(2019)St{\"u}tzle and L{\'o}pez-Ib{\'a}{\~n}ez]{stutzle2019automated}
St{\"u}tzle, T. and L{\'o}pez-Ib{\'a}{\~n}ez, M.
\newblock Automated design of metaheuristic algorithms.
\newblock \emph{Handbook of metaheuristics}, pp.\  541--579, 2019.

\bibitem[Sui et~al.(2023)Sui, Ding, Xia, Liu, and Bu]{sui2023neuralgls}
Sui, J., Ding, S., Xia, B., Liu, R., and Bu, D.
\newblock Neuralgls: learning to guide local search with graph convolutional network for the traveling salesman problem.
\newblock \emph{Neural Computing and Applications}, pp.\  1--20, 2023.

\bibitem[Taillard(1993)]{taillard1993benchmarks}
Taillard, E.
\newblock Benchmarks for basic scheduling problems.
\newblock \emph{european journal of operational research}, 64\penalty0 (2):\penalty0 278--285, 1993.

\bibitem[Van~Laarhoven et~al.(1987)Van~Laarhoven, Aarts, van Laarhoven, and Aarts]{van1987simulated}
Van~Laarhoven, P.~J., Aarts, E.~H., van Laarhoven, P.~J., and Aarts, E.~H.
\newblock \emph{Simulated annealing}.
\newblock Springer, 1987.

\bibitem[Vinyals et~al.(2015)Vinyals, Fortunato, and Jaitly]{vinyals2015pointer}
Vinyals, O., Fortunato, M., and Jaitly, N.
\newblock Pointer networks.
\newblock \emph{Advances in neural information processing systems}, 28, 2015.

\bibitem[Voudouris \& Tsang(1999)Voudouris and Tsang]{voudouris1999guided}
Voudouris, C. and Tsang, E.
\newblock Guided local search and its application to the traveling salesman problem.
\newblock \emph{European journal of operational research}, 113\penalty0 (2):\penalty0 469--499, 1999.

\bibitem[Wei et~al.(2022)Wei, Wang, Schuurmans, Bosma, Xia, Chi, Le, Zhou, et~al.]{wei2022chain}
Wei, J., Wang, X., Schuurmans, D., Bosma, M., Xia, F., Chi, E., Le, Q.~V., Zhou, D., et~al.
\newblock Chain-of-thought prompting elicits reasoning in large language models.
\newblock \emph{Advances in Neural Information Processing Systems}, 35:\penalty0 24824--24837, 2022.

\bibitem[Wu et~al.(2023)Wu, Zhong, Wu, and Tan]{wu2023llm}
Wu, X., Zhong, Y., Wu, J., and Tan, K.~C.
\newblock As-llm: When algorithm selection meets large language model.
\newblock \emph{arXiv preprint arXiv:2311.13184}, 2023.

\bibitem[Wu et~al.(2024)Wu, Wu, Wu, Feng, and Tan]{wu2024evolutionary}
Wu, X., Wu, S.-h., Wu, J., Feng, L., and Tan, K.~C.
\newblock Evolutionary computation in the era of large language model: Survey and roadmap.
\newblock \emph{arXiv preprint arXiv:2401.10034}, 2024.

\bibitem[Xiao \& Wang(2023)Xiao and Wang]{xiao2023llm}
Xiao, H. and Wang, P.
\newblock Llm a*: Human in the loop large language models enabled a* search for robotics.
\newblock \emph{arXiv preprint arXiv:2312.01797}, 2023.

\bibitem[Xin et~al.(2020)Xin, Song, Cao, and Zhang]{xin2020step}
Xin, L., Song, W., Cao, Z., and Zhang, J.
\newblock Step-wise deep learning models for solving routing problems.
\newblock \emph{IEEE Transactions on Industrial Informatics}, 17\penalty0 (7):\penalty0 4861--4871, 2020.

\bibitem[Xu et~al.(2023{\natexlab{a}})Xu, Sun, Zheng, Geng, Zhao, Feng, Tao, and Jiang]{xu2023wizardlm}
Xu, C., Sun, Q., Zheng, K., Geng, X., Zhao, P., Feng, J., Tao, C., and Jiang, D.
\newblock Wizardlm: Empowering large language models to follow complex instructions.
\newblock \emph{arXiv preprint arXiv:2304.12244}, 2023{\natexlab{a}}.

\bibitem[Xu et~al.(2023{\natexlab{b}})Xu, Banburski-Fahey, and Jojic]{xu2023reprompting}
Xu, W., Banburski-Fahey, A., and Jojic, N.
\newblock Reprompting: Automated chain-of-thought prompt inference through gibbs sampling.
\newblock \emph{arXiv preprint arXiv:2305.09993}, 2023{\natexlab{b}}.

\bibitem[Yang et~al.(2023)Yang, Wang, Lu, Liu, Le, Zhou, and Chen]{yang2023large}
Yang, C., Wang, X., Lu, Y., Liu, H., Le, Q.~V., Zhou, D., and Chen, X.
\newblock Large language models as optimizers.
\newblock \emph{arXiv preprint arXiv:2309.03409}, 2023.

\bibitem[Yao et~al.(2023)Yao, Li, and Zhao]{yao2023beyond}
Yao, Y., Li, Z., and Zhao, H.
\newblock Beyond chain-of-thought, effective graph-of-thought reasoning in large language models.
\newblock \emph{arXiv preprint arXiv:2305.16582}, 2023.

\bibitem[Zhang et~al.(2023)Zhang, Mei, Nguyen, and Zhang]{zhang2023survey}
Zhang, F., Mei, Y., Nguyen, S., and Zhang, M.
\newblock Survey on genetic programming and machine learning techniques for heuristic design in job shop scheduling.
\newblock \emph{IEEE Transactions on Evolutionary Computation}, 2023.

\bibitem[Zhao et~al.(2023)Zhao, Lee, and Hsu]{zhao2023large}
Zhao, Z., Lee, W.~S., and Hsu, D.
\newblock Large language models as commonsense knowledge for large-scale task planning.
\newblock \emph{arXiv preprint arXiv:2305.14078}, 2023.

\end{thebibliography}
\bibliographystyle{icml2024}

\newpage
\appendix
\onecolumn

\definecolor{darkbrown}{HTML}{8B4513}
\definecolor{navyblue}{HTML}{000080}
\definecolor{darkgreen}{HTML}{008000}
\definecolor{purple}{HTML}{800080}
\definecolor{red}{HTML}{FF0000}

\section{Additional Related Work}

\subsection{Neural Solvers}
Recently, much effort has been made to develop end-to-end neural solvers, especially for combinatorial optimization~\cite{bengio2021machine}. A pointer network as a neural solver was proposed to construct solutions autoregressively in~\citet{vinyals2015pointer}. Some improvements to this work have been made including using an efficient reinforcement learning framework to replace the original supervised learning~\cite{bello2016neural} and adopting an attention model to replace the pointer network~\cite{kool2018attention,deudon2018learning}. Among these variants~\cite{xin2020step,kwon2020pomo,kwon2021matrix}, a multiple optimal policy proposed in~\cite{kwon2020pomo} archives the state-of-the-art performance on diverse problems of small or moderate sizes. Recent progress, leveraging a heavy encoder and light decoder, enhances its scalarization~\cite{drakulic2023bq,luo2023neural}.

Some effort has also been made to hybridize neural solvers and classic heuristic algorithms including heat map-based methods~\cite{joshi2019efficient,fu2021generalize, qiu2022DIMES}, heat map guided beam search~\citep{joshi2019efficient}, Monte Carlo tree search (MCTS)~\cite{fu2021generalize}, dynamic programming~\cite{kool2022deep}, and guided local search~\cite{hudson2021graph,sui2023neuralgls}. 
The hybridization often requires significant effort to design and train domain neural models.

\subsection{Prompt Engineering}
LLMs with simple promoting face challenges in solving complex reasoning tasks~\cite{chu2023survey}. To address this challenge, Chain-of-Thought (CoT)~\cite{wei2022chain} was proposed to do in-context learning with step-by-step reasoning processes to enhance the reasoning ability of LLMs without human annotation. 

Consequently, there have been many extensions and modifications on CoT such as multi-sampling, tree-of-thoughts~\cite{long2023large}, graph-of-thoughts~\cite{yao2023beyond}, algorithm-of-thoughts~\cite{sel2023algorithm}, and automatic CoT construction~\cite{kojima2022large,xu2023reprompting}. The prompt strategies used in EoH can be regarded as variants of CoT for the design of heuristics with parent heuristics and instructions as in-context information.

\section{Online Bin Packing Problem}

\subsection{Prompt Engineering}
In the following, we introduce the detailed prompt engineering used for online bin packing problem. The prompt engineering for each evolution procedure consists of five main parts: 1) Task description, 2) Strategy-specific prompt, 3) Expected output, 4) Note, and  5) Parent heuristic(s). Figure~\ref{fig:prompt_bp} presents two examples of prompts: initialization prompt and E2 prompt. The prompt engineering for other operators (E1, M1, M2, and S1) has the same structure as these two examples and is not listed for brevity. The five parts in different colors are introduced as follows:

\begin{itemize}[itemsep=-3pt]
    \item \textcolor {darkbrown}{\textbf{Task description}}: It informs LLMs of the problem description. Different prompt strategies usually share the same task description.
    \item {\textbf{Strategy-specific prompt}}: It instructs LLMs to do reasoning over the in-context information and generate new heuristics as well as its corresponding code implementation. Different prompt strategies have different strategy-specific prompts. For example, during initialization, we instruct the LLM to create a totally new heuristic. During evolution, we request LLM to perform different types of reasoning over parent heuristic(s) to explore the heuristic search space.
    \item \textcolor{navyblue}{\textbf{Expected output}}: It asks the LLM to provide a description of the designed heuristic and then produce the code implementation for the heuristic. In this paper, the code implementation is a function in Python. We explicitly define the name, input, and output of the code for easy identification by \ourmethod{} framework. 
    \item \textcolor{purple}{\textbf{Note}}: It provides additional instructions for LLM to improve the efficiency and robustness. For example, we may suggest specific types of inputs and outputs and discourage extra explanations to prevent a long response.
    \item \textcolor{darkgreen}{\textbf{Parent heuristic(s)}}: It includes parent heuristic(s) to enable in-context learning over both the linguistic heuristic description and the code implementation. The initialization prompt does not include this part.
\end{itemize}

\begin{figure}
    \centering
    \caption{Two examples of prompt engineering used in initialization prompt and E2 prompt for online bin packing.}
    \label{fig:prompt_bp}

\begin{minipage}{0.38\textwidth}
\small
\begin{dialogbox}
  \textbf{Prompt for Initialization} \\
  
  \textcolor {darkbrown}{ I need help designing a new heuristic that scores a set of bins to assign an item. In each step, the item will be assigned to the bin with the maximum score. If the rest capacity of a bin equals the maximum capacity, it will not be used. The final goal is to minimize the number of used bins.} \\

  Please design a new heuristic.
  
  \textbf{Firstly,} describe your \textcolor{navyblue}{new heuristic and main steps in one sentence}. \\
  \textbf{Next,} implement it in Python as a function named \textcolor{navyblue}{'score'}. This function should accept \textcolor{navyblue}{two inputs: 'item' and 'bins'}. The function should return \textcolor{navyblue}{one output: 'scores'}. \textcolor{navyblue}{'item' and 'bins' are the size of the current item and the rest capacities of feasible bins, which are larger than the item size. The output named 'scores' is the scores for the bins for assignment.}  \\

  \textcolor{purple}{Note that 'item' is of type int, 'bins' is a Numpy array that includes integer values, and 'scores' should be a Numpy array. Avoid utilizing the random component, and it is crucial to maintain self-consistency. Do not give additional explanations.}
\end{dialogbox}
\end{minipage}
\begin{minipage}{0.6\textwidth}
\small
\begin{dialogbox}
  \textbf{Prompt for E2} \\
  
  \textcolor[HTML]{8B4513}{ I need help designing a new heuristic that scores a set of bins to assign an item. In each step, the item will be assigned to the bin with the maximum score. If the rest capacity of a bin equals the maximum capacity, it will not be used. The final goal is to minimize the number of used bins.} \\

  \textcolor{darkgreen}{I have five existing heuristics with their codes as follows: \\
  No.1 Heuristic description: \\
       Code:   \\
       ... \\
  No.5 Heuristic description: \\
       Code: \\  } 
       
   Please help me design a new heuristic that is different from the given ones but can be motivated by them. \\
   \textbf{Firstly,} identify the common idea in the provided heuristics.   \\
   \textbf{Secondly,} based on the backbone idea \textcolor{navyblue}{describe your new heuristic in one sentence.} \\
   \textbf{Thirdly,} implement it in Python as a function named \textcolor{navyblue}{'score'}. This function should accept \textcolor{navyblue}{two inputs: 'item' and 'bins'}. The function should return \textcolor{navyblue}{one output: 'scores'}. \textcolor{navyblue}{'item' and 'bins' are the size of the current item and the rest capacities of feasible bins, which are larger than the item size. The output named 'scores' is the scores for the bins for assignment.}  \\

  \textcolor{purple}{Note that 'item' is of type int, 'bins' is a Numpy array that includes integer values, and 'scores' should be a Numpy array. Avoid utilizing the random component, and it is crucial to maintain self-consistency. Do not give additional explanations.}
\end{dialogbox}
\end{minipage}
\end{figure}

\begin{figure}[htbp]
    \centering
    \includegraphics[width=\textwidth]{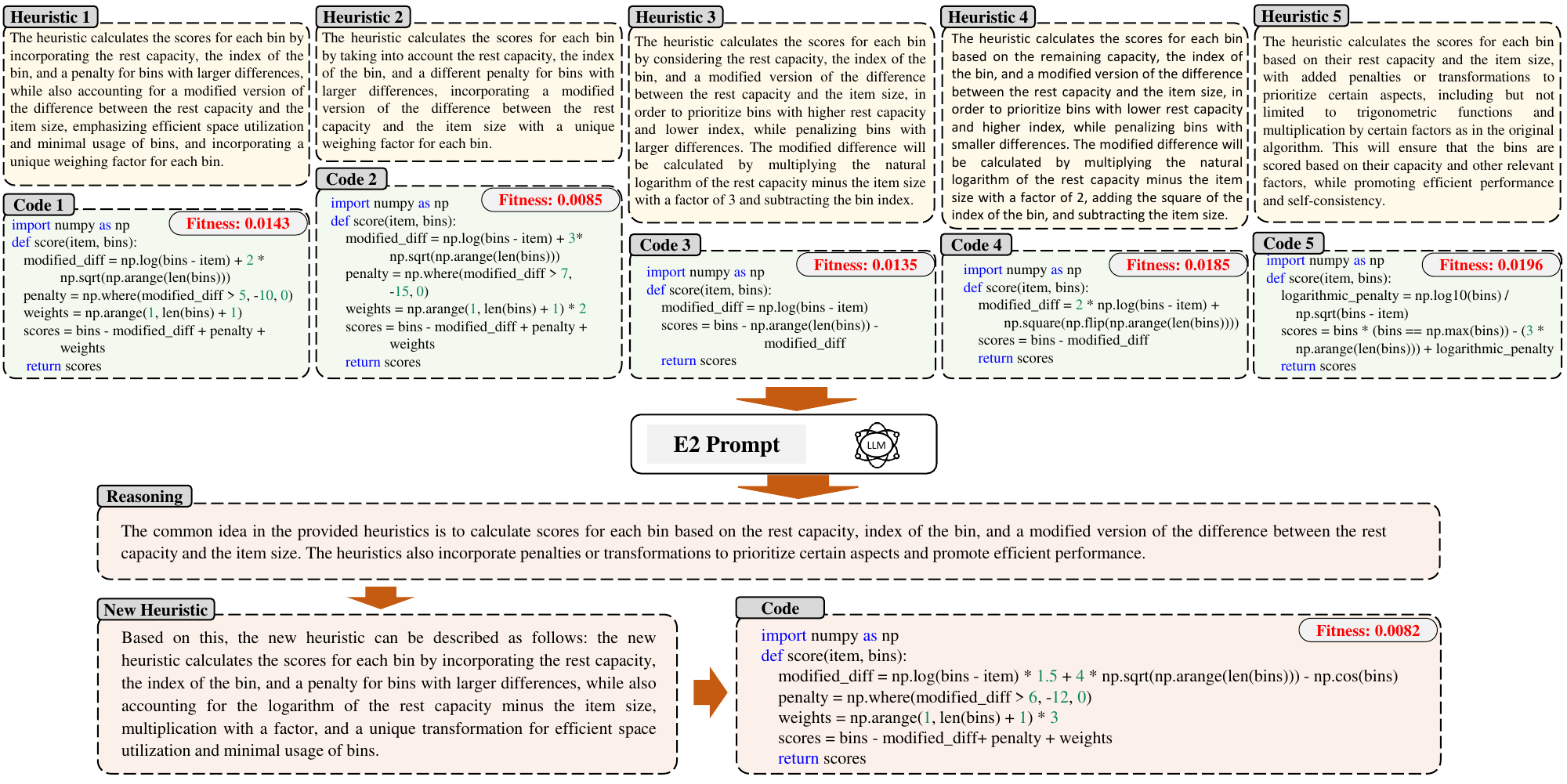}~\label{fig:ee2}
    \caption{Illustration of the generation of new heuristic and its code implementation using E2 prompt in one step in \ourmethod{} on online bin packing problem. Five parent heuristics are selected from the population. E2 prompts LLM to first observe and summarize the common idea in the five heuristics, then design a new heuristic using this idea, and finally give a code implementation for the designed heuristic.}
    \label{fig:enter-label}
\end{figure}

\subsection{Heuristic Evolution}~\label{e2}


The convergence process of evolution has been discussed in the main part. Figure 6 presents a detailed illustration of  E2 in one generation. Five heuristics are chosen from the population, each of them consists of a high-level description and a detailed Python implementation. These heuristics are then used as inputs for prompt engineering in E2, which was introduced in the previous section. The LLMs are given three instructions: firstly, to identify the shared concept among the provided heuristics based on their descriptions and codes; secondly, to design a new heuristic based on this analysis; and thirdly, to implement the heuristic in Python with a given name, input, and output. The improved heuristic follows the identified common idea and integrates new components.

\subsection{Designed Heuristic}
Comparison of heuristics designed by humans, EoC, FunSearch, and our proposed \ourmethod{} is presented in Figure~\ref{fig:heuristics_bp}. We present them as Python functions with the same inputs and output. The inputs 'item' and 'bins' represent the size of the current item and the rest capacities of bins following~\citet{romera2024mathematical}. The output is the scores assigned to the bins. At each step, the item is assigned to the bin with the highest score. It is worth noting that the two most commonly utilized hand-crafted heuristics can be implemented in one line of code. On the other hand, the heuristics designed by EoC, FunSearch, and our method \ourmethod{} are more complicated, making them difficult to achieve for human designers.

\begin{figure}[htbp]
\caption{Heuristics produced using different approaches: First fit and Best fit by human designers. Evolution-of-Codes (EoC), FunSearch, and our proposed Evolution-of-Heuristics (\ourmethod{}). }~\label{fig:heuristics_bp}
\begin{minipage}{0.48\linewidth}
\begin{lstlisting}

|\textbf{Heuristic Designed by \textcolor{red}{Human (First Fit)}}|

import numpy as np
def heuristic(item, bins):
    scores = -np.arange(len(bins))
    return scores
    
\end{lstlisting}
\end{minipage}\hfill
\begin{minipage}{0.48\linewidth}
\begin{lstlisting}

|\textbf{Heuristic Designed by \textcolor{red}{Human (Best Fit)}}|

def heuristic(item, bins):
    scores = item - bins
    return scores

\end{lstlisting}
\end{minipage}\hfill
\begin{minipage}{0.48\linewidth}
\begin{lstlisting}

|\textbf{Heuristic Designed by \textcolor{red}{EoC}}|

import numpy as np
def heuristic(item, bins):
    scores = np.log(item) * (bins ** 2) / (item * np.sqrt(bins - item)) + (bins / item) ** 3
    scores[bins == bins.max()] = -np.inf
    return scores

\end{lstlisting}
\end{minipage}\hfill
\begin{minipage}{0.48\linewidth}
\begin{lstlisting}

|\textbf{Heuristic Designed by \textcolor{red}{FunSearch}}|

def heuristic(item, bins):
  max_bin_cap = max(bins)
  score = (bins - max_bin_cap)**2 / item + bins**2 / (item**2)
  score += bins**2 / item**3
  score[bins > item] = -score[bins > item]
  score[1:] -= score[:-1]
  return score
  
\end{lstlisting}
\end{minipage}\hfill

\begin{minipage}{0.99\linewidth}
\begin{lstlisting}

|\textbf{Heuristic Designed by \textcolor{red}{\ourmethod{}}}|

|\textbf{<Heuristic Description>}|
|\text{The heuristic incorporates a weighted average of the utilization ratio,}|
|\text{dynamic adjustment, and an exponentially decaying factor, with different}|
|\text{parameter settings to minimize the number of used bins.}|

|\textbf{<Code>}|
import numpy as np
def heuristic(item, bins):
  diff = bins-item  # remaining capacity
  exp = np.exp(diff)  # exponent term
  sqrt = np.sqrt(diff)  # square root term
  ulti = 1-diff/bins  # utilization term
  comb = ulti * sqrt  # combination of utilization and square root 
  adjust = np.where(diff > (item * 3), comb + 0.8, comb + 0.3)
      # hybrid adjustment term to penalize large bins 
  hybrid_exp = bins / ((exp + 0.7) *exp)
      # hybrid score based on exponent term
  scores = hybrid_exp + adjust
      # sum of hybrid score and adjustment
  return scores
\end{lstlisting}
\end{minipage}
\end{figure}

\subsection{More Results}~\label{more_results_bp}

Table~\ref{table:weibull_appendix} summarizes the results of First Fit, Best Fit, and the algorithm produced by FunSearch, EoC, and EoH on Weibull instances with different capacities and sizes. The average gap to the lower bound on five instances is reported for different heuristics. The best results, indicated in bold, represent the heuristics with the lowest average gap to the lower bound. We observe that the heuristic designed by EoC overfits on the distribution (100, 5k) which is used to evaluate the fitness of heuristics generated in the evolution process. Among all the heuristics, EoH consistently achieves the best performance, with an average gap to the lower bound of 1.18\%. Importantly, EoH only utilizes a few thousand queries, which accounts for much less computational budget required by FunSearch (around 1 million queries reported in~\citet{romera2024mathematical}). EoH achieves the same best gap in the training distribution while demonstrating superior generalization performance, particularly for different capacities. 

\begin{table}[htbp]
\centering

\caption{Results on Weibull instances with varied capacities and problem sizes. Average gap to lower bound across five instances (best results in bold).}~\label{table:weibull_appendix}
\tiny
\resizebox{0.65\textwidth}{!}{%
\begin{tabular}{ccccclc}
\toprule
Capacity & Size & First Fit & Best Fit & FunSearch & \multicolumn{1}{c}{EoC}     & \ourmethod{}     \\
\midrule
100    & 1k & 5.32\%    & 4.87\%   & 3.78\%  & 148.63\%& \textbf{2.24\%} \\
 & 5k & 4.40\%    & 4.08\%   & \textbf{0.80\%} & 3.23\%& \textbf{0.80\%} \\
 & 10k& 4.44\%    & 4.09\%   & \textbf{0.33\%} & 24.55\% & 0.61\%  \\
 \midrule
200    & 1k & 4.86\%    & 4.42\%   & 4.20\%  & 134.59\%& \textbf{2.10\%} \\
 & 5k & 4.14\%    & 3.80\%   & 0.93\%  & 5.26\%& \textbf{0.76\%} \\
 & 10k& 4.16\%    & 3.83\%   & \textbf{0.36\%} & 23.96\% & 0.58\%  \\
 \midrule
300    & 1k & 4.93\%    & 4.48\%   & 4.93\%  & 141.48\%& \textbf{2.18\%} \\
 & 5k & 4.18\%    & 3.83\%   & 1.07\%  & 8.31\%& \textbf{0.77\%} \\
 & 10k& 4.20\%    & 3.87\%   & \textbf{0.49\%} & 27.62\% & 0.59\%  \\
 \midrule
400    & 1k & 4.97\%    & 4.50\%   & 5.38\%  & 150.06\%& \textbf{2.16\%} \\
 & 5k & 4.24\%    & 3.88\%   & 1.57\%  & 10.52\% & \textbf{0.79\%} \\
 & 10k& 4.25\%    & 3.91\%   & 0.69\%  & 30.45\% & \textbf{0.61\%} \\
 \midrule
500    & 1k & 4.97\%    & 4.50\%   & 6.75\%  & 150.89\%& \textbf{2.13\%} \\
 & 5k & 4.27\%    & 3.91\%   & 1.47\%  & 12.53\% & \textbf{0.78\%} \\
 & 10k& 4.28\%    & 3.95\%   & 0.74\%  & 32.02\% & \textbf{0.61\%} \\
 \midrule
\multicolumn{2}{c}{Average} & 4.51\%    & 4.13\%   & 2.23\%  & \multicolumn{1}{c}{60.27\%} & \textbf{1.18\%}\\
\bottomrule
\end{tabular}%
}
\end{table}

\section{Traveling Salesman Problem}

\subsection{Guided Local Search}

The heuristics designed by \ourmethod{} work with local search operators in a guided local search framework for both TSP and FSSP. In contrast to the complicated local search in SOTA solvers~\cite{LKH3}, We only adopt the two basic local search operators. We show that even with these two simple operators, \ourmethod{} can design a very competitive algorithm.

Guided local search (GLS) is a widely used strategy to guide a local search escape from local optimal solutions for solving combinatorial optimization problems ~\cite{alsheddy2018guided,voudouris1999guided}. In a typical local search~\cite{alsheddy2018guided}, when a local search is trapped in a local optimal solution, GLS modifies the objective function to guide the local search to move to other promising search regions~\cite {alsheddy2018guided}.

A GLS algorithm alternates the two phases: \textit{local search} and \textit{perturbation}~\cite{arnold2019knowledge}. During the \textit{local search} phase, we employ local search operators to search. When the search is trapped in a local optimum, the \textit{perturbation} phase is invoked to update the objective function (i.e., landscape) using a heuristic strategy.

\ourmethod{} is employed to design a strategy to update the objective function. With some pre-selected local search operators, this strategy can define a GLS heuristic. In our experiments, for each updating strategy generated in EoH, the performance of its corresponding GLS heuristic is evaluated on a set of problem instances, with the average performance as its fitness value.

\subsection{Prompt Engineering}

\begin{figure}[htbp]
    \centering
    \caption{Two examples of prompt engineering used in initialization and E2 strategy for TSP.}
    \label{fig:prompt_tsp}

\begin{minipage}{0.38\textwidth}
\small
\begin{dialogbox}
  \textbf{Prompt for Initialization} \\
  
  \textcolor {darkbrown}{ Given an edge distance matrix and a local optimal route, please help me design a strategy to update the distance matrix to avoid being trapped in the local optimum with the final goal of finding a tour with minimized distance. You should create a heuristic for me to update the edge distance matrix.} \\

  Please design a new heuristic.
  
  \textbf{Firstly,} describe your \textcolor{navyblue}{new heuristic and main steps in one sentence}. \\
  \textbf{Next,} implement it in Python as a function named \textcolor{navyblue}{'update\_edge\_distance'}. This function should accept \textcolor{navyblue}{three inputs: 'edge\_distance', 'local\_opt\_tour', and 'edge\_n\_used'}. The function should return \textcolor{navyblue}{one output: 'updated\_edge\_distance'}. \textcolor{navyblue}{'local\_opt\_tour' includes the local optimal tour of IDs, 'edge\_distance' and 'edge\_n\_used' are matrixes, 'edge\_n\_used' includes the number of each edge used during perturbation.}  \\

  \textcolor{purple}{All inputs and outputs are Numpy arrays. Do not give additional explanations.}
\end{dialogbox}
\end{minipage}
\begin{minipage}{0.61\textwidth}
\small
\begin{dialogbox}
  \textbf{Prompt for E2} \\
  
  \textcolor[HTML]{8B4513}{ Given an edge distance matrix and a local optimal route, please help me design a strategy to update the distance matrix to avoid being trapped in the local optimum with the final goal of finding a tour with minimized distance. You should create a heuristic for me to update the edge distance matrix.} \\

  \textcolor{darkgreen}{I have five existing heuristics with their codes as follows: \\
  No.1 Heuristic description: \\
       Code:   \\
       ... \\
  No.5 Heuristic description: \\
       Code: \\  } 
       
   Please help me design a new heuristic that is different from the given ones but can be motivated by them. \\
   \textbf{Firstly,} identify the common idea in the provided heuristics.   \\
   \textbf{Secondly,} based on the backbone idea \textcolor{navyblue}{describe your new heuristic in one sentence.} \\
   \textbf{Thirdly,} implement it in Python as a function named \textcolor{navyblue}{'update\_edge\_distance'}. This function should accept \textcolor{navyblue}{three inputs: 'edge\_distance', 'local\_opt\_tour', and 'edge\_n\_used'}. The function should return \textcolor{navyblue}{one output: 'updated\_edge\_distance'}. \textcolor{navyblue}{'local\_opt\_tour' includes the local optimal tour of IDs, 'edge\_distance' and 'edge\_n\_used' are matrixes, 'edge\_n\_used' includes the number of each edge used during perturbation.}  \\

  \textcolor{purple}{All inputs and outputs are Numpy arrays. Do not give additional explanations.}
\end{dialogbox}
\end{minipage}
\end{figure}

In the following, we give details of prompt engineering used for TSP. We use the same five components outlined in the prompt engineering section for bin packing. In Figure~\ref{fig:prompt_tsp}, we provide two illustrative examples of prompts for initialization and E2, with each component represented in different colors. For the sake of brevity, we omit the prompt engineering details for other prompt strategies.


\subsection{Designed Heuristic Strategy}

Figure~\ref{fig:heuristic_tsp} illustrates the heuristic strategy designed by \ourmethod{} for updating the distance matrix for TSP. This matrix is obtained by modifying the original distance matrix and adding some random factors, which involves several intermediate variables, including the average of distances, the average of edge used, and a penalty on the current local optimal route.

\begin{figure}[ht]
\caption{Heuristic strategy designed by \ourmethod{} on TSP. }~\label{fig:heuristic_tsp}

\begin{minipage}{0.99\linewidth}
\begin{lstlisting}

|\textbf{Heuristic Designed by \textcolor{red}{\ourmethod{}}}|

|\textbf{<Heuristic Description>}|
|\text{Update the edge distances in the edge distance matrix by }|
|\text{incorporating a pheromone-like effect, where the update is determined by edge count, }|
|\text{distance, and usage, with the addition of a decay factor to }|
|\text{avoid stagnation and promote exploration. }|

|\textbf{<Code>}|
import numpy as np

def heuristic(edge_distance, local_opt_tour, edge_n_used):
    updated_edge_distance = np.copy(edge_distance)
    edge_count = np.zeros_like(edge_distance)
    for i in range(len(local_opt_tour) - 1):
        start = local_opt_tour[i]
        end = local_opt_tour[i + 1]
        edge_count[start][end] += 1
        edge_count[end][start] += 1
        # penalize local optimal route
    edge_n_used_max = np.max(edge_n_used) 
        # calculate the average edge used
    decay_factor = 0.1 # decay fastor
    mean_distance = np.mean(edge_distance) 
        # calculate the average distance
    for i in range(edge_distance.shape[0]):
        for j in range(edge_distance.shape[1]):
            if edge_count[i][j] > 0:
                noise_factor = (np.random.uniform(0.7, 1.3) / edge_count[i][j]) + (edge_distance[i][j] / mean_distance) - (0.3 / edge_n_used_max) * edge_n_used[i][j]
                    # calculate a hybrid noise factor
                updated_edge_distance[i][j] += noise_factor * (1 + edge_count[i][j]) - decay_factor * updated_edge_distance[i][j]
                    # The new guiding edge distance matrix is calculated based on both a noise term and a decayed original distance matrix

    return updated_edge_distance
\end{lstlisting}
\end{minipage}
\end{figure}

\subsection{More Experimental Results}~\label{more_results_tsp}

We also compare the heuristic generated by EoH with the following methods: 
\begin{itemize}
    \item Graph Convolutional Network (GCN) method for TSP~\cite{joshi2019efficient}. 
    \item  Attention Model (AM)~\cite{kool2018attention}. It uses neural networks to learn heuristics for combinatorial optimization. 
    \item  POMO~\cite{kwon2020pomo}. It adopts AM ideas and achieves state-of-the-art results. 
    \item LEHD~\cite{luo2023neural}. It is a new variant of AM with a different heavy decoder structure and is trained using supervised learning. 
    \item GLS~\cite{voudouris1999guided}. It is the vanilla version of GLS for TSP.
    \item EBGLS~\cite{shi2018eb}. It extends the GLS by considering the big valley feature of the TSP. 
    \item KGLS~\cite{arnold2019knowledge}. It uses multiple features extracted from previous knowledge of routing problems.  
    \item GNNGLS~\cite{hudson2021graph} and  NeuralGLS~\cite{sui2023neuralgls}. They use deep learning models in GLS. 
\end{itemize}
We set the maximum number of calls of LS to be 1,000 for each GLS algorithm on every test instance. We use the source code of POMO~\cite{kwon2020pomo} , BQ~\cite{drakulic2023bq}, and LEHD~\cite{luo2023neural} in our experiments. The experimental results for GNNGLS~\citet{hudson2021graph}, NeuralGLS~\citet{sui2023neuralgls}, AM~\citet{hudson2021graph} and GCN ~\citet{sui2023neuralgls} are directly extracted from their respective papers. The solutions obtained by Concorde~\cite{applegate2006concorde} are used as the baselines for computing the performance gap. 

We consider three different numbers of locations: 20, 50, and 100. For each number of locations, we randomly generate 1,000 locations from $[0,1]^2$ and thus obtain 1,000 test instances.  
Table~\ref{table:tsprandom_appendix} shows the average performance of the heuristics on these random instances. The table provides the average gap compared with the baseline solver Concorde, as well as the average running time on each instance. It should be pointed out that POMO, BQ, and LEHD run in a parallel manner on the GPU, so single-instance running time is not provided. It is very clear from Table~\ref{table:tsprandom_appendix} that the heuristic produced by 
EoH performs the best.  



\begin{table}[tbp]
\centering
\caption{Results on TSP20, TSP50, and TSP100. The gap and time are averaged over 1,000 instances.}~\label{table:tsprandom_appendix}
\small
\resizebox{0.7\textwidth}{!}{%
\begin{tabular}{lcccccc}
\toprule
\multirow{2}{*}{Method}& \multicolumn{2}{c}{TSP20}   & \multicolumn{2}{c}{TSP50}   & \multicolumn{2}{c}{TSP100}   \\

& Gap (\%) & Time (s) & Gap (\%) & Time (s) & Gap (\%) & Time (s) \\
\midrule
 Concorde & 0.000    & 0.010    & 0.000    & 0.051    & 0.000    & 0.224    \\
 LKH3     & 0.000    & 0.020    & 0.000    & 0.069    & 0.011    & 0.118    \\
\midrule
NN  & 17.448   & 0.000    & 23.230   & 0.002    & 25.104   & 0.010    \\
 FI & 2.242    & 0.005    & 7.263    & 0.065    & 12.456   & 0.444    \\
 AM & 0.069    & 0.038    & 0.494    & 0.124    & 2.368    & 0.356    \\
 GCN& 0.035    & 0.974    & 0.884    & 3.080    & 1.880    & 6.127    \\
 POMO   & 0.120    & /  & 0.640    & /  & 1.070    & /  \\
 POMO aug8& 0.000    & /  & 0.030    & /  & 0.140    & /  \\
 BQ & 0.379    & /  & 0.245    & /  & 0.579    & /  \\
 LEHD     & 0.950    & /  & 0.485    & /  & 0.577    & /  \\
\midrule
 LS & 1.814    & 0.006    & 3.461    & 0.006    & 4.004    & 0.008    \\
GLS& 0.004    & 0.088    & 0.045    & 0.248    & 0.659    & 0.683    \\
 EBGLS    & 0.002    & 0.091    & 0.003    & 0.276    & 0.155    & 0.779    \\
 KGLS     & \textbf{0.000}    & 1.112    & \textbf{0.000}    & 3.215    & 0.035    & 7.468    \\
 GNNGLS & \textbf{0.000}    & 10.010   & 0.009    & 10.037   & 0.698    & 10.108   \\
 NeuralGLS & \textbf{0.000}    & 10.005   & 0.003    & 10.011   & 0.470    & 10.024   \\
\midrule
\ourmethod{}  & \textbf{0.000}    & 0.498    & \textbf{0.000}    & 1.494    & \textbf{0.025}    & 4.510   \\
\bottomrule
\end{tabular}%
}
\end{table}

We also conduct experiments on 29 TSPLib instances.
As shown in Table~\ref{table:TSPLibresults}, The GLS algorithm designed by \ourmethod{} outperforms all the other heuristics including hand-crafted ones in terms of the average gap on the 29 instances. 


\begin{table*}[htbp]
\centering
\caption{Results on TSPLib instances. The gap (\%) to the best-known solution from TSPLib.}~\label{table:TSPLibresults}
\large
\renewcommand\arraystretch{0.95}
\resizebox{0.7\textwidth}{!}{%
\begin{tabular}{ccccccccccc}
\toprule
\multicolumn{1}{c}{\multirow{2}{*}{Instance}} & \multicolumn{3}{c}{Other Algorithms} & \multicolumn{6}{c}{GLS Algorithms} & \multirow{2}{*}{\ourmethod{}} \\
\multicolumn{1}{c}{}  & AM& POMO & LEHD& GNNGLS & NeuralGLS & LS    & GLS   & EBGLS & KGLS  &  \\
\midrule
eil51    & 1.63   & 0.83  & 1.64  & 0.00 & 0.00 & 2.85 & 0.67 & 0.67 & 0.67 & 0.67 \\
berlin52 & 4.17   & 0.04  & 0.03  & 0.14 & 0.00 & 3.89 & 0.03 & 0.03 & 0.03 & 0.03 \\
st70     & 1.74   & 0.31  & 0.33  & 0.76 & 0.00 & 2.64 & 0.31 & 0.31 & 0.31 & 0.31 \\
eil76    & 1.99   & 1.18  & 2.54  & 0.16 & 0.00 & 3.93 & 1.37 & 1.18 & 1.18 & 1.48 \\
pr76     & 0.82   & 0.00  & 0.22  & 0.04 & 0.82 & 6.71 & 0.00 & 0.00 & 0.00 & 0.00 \\
rat99    & 2.65   & 2.39  & 1.10  & 0.55 & 0.72 & 6.58 & 1.55 & 0.74 & 0.68 & 0.68 \\
kroA100  & 4.02   & 0.41  & 0.12  & 0.73 & 0.03 & 3.00 & 0.02 & 0.02 & 0.06 & 0.02 \\
kroB100  & 5.14   & 0.32  & 0.26  & 0.15 & 0.88 & 0.58 & 0.23 & 0.00 & 0.25 & 0.00 \\
kroC100  & 0.97   & 0.18  & 0.32  & 1.57 & 1.77 & 4.70 & 0.50 & 0.01 & 0.01 & 0.01 \\
kroD100  & 2.72   & 0.84  & 0.38  & 0.57 & 0.00 & 5.67 & 0.00 & 0.20 & 0.00 & 0.00 \\
kroE100  & 1.47   & 0.45  & 0.43  & 1.22 & 1.05 & 4.64 & 0.49 & 0.00 & 0.07 & 0.14 \\
rd100    & 3.41   & 0.01  & 0.01  & 0.46 & 0.00 & 1.27 & 0.01 & 0.01 & 0.02 & 0.01 \\
eil101   & 2.99   & 1.84  & 2.31  & 0.20 & 0.36 & 8.82 & 3.28 & 1.91 & 2.07 & 2.27 \\
lin105   & 1.74   & 0.52  & 0.34  & 0.61 & 0.65 & 1.87 & 0.03 & 0.03 & 0.03 & 0.03 \\
pr107    & 3.93   & 0.52  & 11.24 & 0.44 & 0.81 & 0.72 & 0.40 & 0.00 & 0.00 & 0.00 \\
pr124    & 3.68   & 0.60  & 1.11  & 0.76 & 0.08 & 2.44 & 0.60 & 0.60 & 0.08 & 0.00 \\
bier127  & 5.91   & 13.72 & 4.76  & 1.95 & 2.73 & 1.79 & 0.59 & 0.29 & 0.42 & 0.42 \\
ch130    & 3.18   & 0.16  & 0.55  & 3.52 & 1.19 & 7.61 & 1.09 & 0.46 & 0.01 & 0.01 \\
pr136    & 5.06   & 0.93  & 0.45  & 3.39 & 2.32 & 6.30 & 2.01 & 0.28 & 0.24 & 0.00 \\
pr144    & 7.64   & 0.53  & 0.19  & 3.58 & 0.74 & 4.19 & 0.09 & 0.00 & 0.00 & 0.00 \\
ch150    & 4.58   & 0.53  & 0.52  & 2.11 & 2.49 & 1.35 & 0.68 & 0.37 & 0.04 & 0.24 \\
kroA150  & 3.78   & 0.70  & 1.40  & 2.98 & 0.77 & 5.05 & 1.75 & 0.26 & 0.17 & 0.00 \\
kroB150  & 2.44   & 1.17  & 0.76  & 3.26 & 3.11 & 5.55 & 1.01 & 0.00 & 0.08 & 0.00 \\
pr152    & 7.49   & 1.05  & 12.14 & 3.12 & 0.00 & 2.75 & 0.19 & 0.19 & 0.19 & 0.19 \\
u159     & 7.55   & 0.95  & 1.13  & 1.02 & 0.90 & 5.63 & 0.74 & 0.78 & 0.96 & 0.00 \\
rat195   & 6.89   & 8.15  & 1.42  & 1.67 & 0.48 & 2.14 & 0.61 & 0.61 & 0.97 & 0.82 \\
d198     & 373.02 & 17.29 & 9.23  & 4.77 & 1.28 & 7.96 & 2.08 & 1.87 & 0.31 & 0.59 \\
kroA200  & 7.11   & 1.58  & 0.64  & 2.03 & 0.86 & 0.91 & 0.75 & 0.18 & 0.71 & 0.15 \\
kroB200  & 8.54   & 1.44  & 0.16  & 2.59 & 3.74 & 4.71 & 1.43 & 1.27 & 0.89 & 0.20 \\
\midrule
Average  & 16.77  & 2.02  & 1.92  & 1.53 & 0.96 & 4.01 & 0.78 & 0.42 & 0.36 & 0.28 \\
\bottomrule
\end{tabular}%
}
\end{table*}

\section{Flow Shop Scheduling Problem}

\subsection{Prompt Engineering}

This subsection is to introduce details of prompt engineering used for FSSP. We use the same components for bin packing. Figure~\ref{fig:prompt_fssp} provides two illustrative examples of prompts for initialization and E2, with each component represented in a different color. 

\begin{figure}[t]
    \centering
    \caption{Two examples of prompt engineering used in initialization and E2 strategy for FSSP.}
    \label{fig:prompt_fssp}

\begin{minipage}{0.38\textwidth}
\small
\begin{dialogbox}
  \textbf{Prompt for Initialization} \\
  
  \textcolor {darkbrown}{I have n jobs and m machines. Help me create a new heuristic to update the execution time matrix and select the top jobs to perturb to avoid being trapped in the local optimum scheduling with the final goal of finding scheduling with minimized makespan.} \\

  Please design a new heuristic.
  
  \textbf{Firstly,} describe your \textcolor{navyblue}{new heuristic and main steps in one sentence}. \\
  \textbf{Next,} implement it in Python as a function named \textcolor{navyblue}{'get\_matrix\_and\_jobs'}. This function should accept \textcolor{navyblue}{four inputs: "current\_sequence","time\_matrix","m","n"}. The function should return \textcolor{navyblue}{two output: "new\_matrix",'perturb\_jobs'}. \textcolor{navyblue}{The variable 'current\_sequence' represents the current sequence of jobs. The variables 'm' and 'n' denote the number of machines and number of jobs, respectively. The variable 'time\_matrix' is a matrix of size n*m that contains the execution time of each job on each machine. The output 'new\_matrix' is the updated time matrix, and 'perturb\_jobs' includes the top jobs to be perturbed.}  \\

  \textcolor{purple}{The matrix and job list are Numpy arrays. Do not give additional explanations.}
\end{dialogbox}
\end{minipage}
\begin{minipage}{0.61\textwidth}
\small
\begin{dialogbox}
  \textbf{Prompt for E2} \\
  
  \textcolor[HTML]{8B4513}{I have n jobs and m machines. Help me create a new heuristic to update the execution time matrix and select the top jobs to perturb to avoid being trapped in the local optimum scheduling with the final goal of finding scheduling with minimized makespan.} \\

  \textcolor{darkgreen}{I have five existing heuristics with their codes as follows: \\
  No.1 Heuristic description: \\
       Code:   \\
       ... \\
  No.5 Heuristic description: \\
       Code: \\  } 
       
   Please help me design a new heuristic that is different from the given ones but can be motivated by them. \\
   \textbf{Firstly,} identify the common idea in the provided heuristics.   \\
   \textbf{Secondly,} based on the backbone idea \textcolor{navyblue}{describe your new heuristic in one sentence.} \\
   \textbf{Thirdly,} implement it in Python as a function named \textcolor{navyblue}{'get\_matrix\_and\_jobs'}. This function should accept \textcolor{navyblue}{four inputs: "current\_sequence","time\_matrix","m","n"}. The function should return \textcolor{navyblue}{two output: "new\_matrix",'perturb\_jobs'}. \textcolor{navyblue}{The variable 'current\_sequence' represents the current sequence of jobs. The variables 'm' and 'n' denote the number of machines and number of jobs, respectively. The variable 'time\_matrix' is a matrix of size n*m that contains the execution time of each job on each machine. The output 'new\_matrix' is the updated time matrix, and 'perturb\_jobs' includes the top jobs to be perturbed.}   \\

  \textcolor{purple}{The matrix and job list are Numpy arrays. Do not give additional explanations.}
\end{dialogbox}
\end{minipage}
\end{figure}

\subsection{Designed Heuristic Strategy}
We adopt the same GLS framework with two selected local search operators: Swap and Relocate. EoH is used to design a heuristic strategy for updating both the execution time matrix and to determine the perturbed jobs. 
Figure~\ref{fig:heuristic_fssp} illustrates a heuristic designed by \ourmethod{} for FSSP. Some selected elements in the original matrix are updated. 

\begin{figure}[ht]
\caption{Heuristic strategy designed by \ourmethod{} on FSSP. }~\label{fig:heuristic_fssp}

\begin{minipage}{0.99\linewidth}
\begin{lstlisting}

|\textbf{Heuristic Designed by \textcolor{red}{\ourmethod{}}}|

|\textbf{<Heuristic Description>}|
|\text{The heuristic randomly selects a subset of machines, computes the weighted }|
|\text{average execution time  for each job on the selected machines, }|
|\text{and perturbs the top jobs in the current sequence to update the }|
|\text{execution time matrix by scaling the original execution time with }|
|\text{a random perturbation factor between 0.8 and 1.2.}|


|\textbf{<Code>}|
import numpy as np
def heuristic(current_sequence, time_matrix, m, n):
    machine_subset = np.random.choice(m, max(1, int(0.3*m)), replace=False)
        # randomly select a subset of machines
    weighted_avg_execution_time = np.average(time_matrix[:, machine_subset], axis=1, weights=np.random.rand(len(machine_subset)))
        # compute the weighted average execution time
    perturb_jobs = np.argsort(weighted_avg_execution_time)[-int(0.3*n):]
        # sort the last jobs based on the weighted average execution time
    new_matrix = time_matrix.copy()
    perturbation_factors = np.random.uniform(0.8, 1.2, size=(len(perturb_jobs), len(machine_subset)))
        # calculate perturbation factors, introduce certain randomness
    new_matrix[perturb_jobs[:, np.newaxis], machine_subset] *= perturbation_factors
        # calculate the final guiding matrix
    return new_matrix, perturb_jobs
\end{lstlisting}
\end{minipage}
\end{figure}

\subsection{More Results}~\label{more_results_fssp}
We also compare the heuristic produced by EoH with the following methods. 
\begin{itemize}
\item GUPTA~\cite{gupta1971functional}. It is a classic heuristic algorithm for FSSP.
\item CDS~\cite{campbell1970heuristic}. It is a classic heuristic algorithm for FSSP.
\item  NEH~\cite{nawaz1983heuristic}. It is widely recognized as an efficient heuristic for FSSP.
\item NEHFF~\cite{fernandez2014insertion}. It is a revision of NEH.
\item PFSPNet and PFSPNet\_NEH~\cite{pan2021deep}. They are recently developed deep learning solvers for flow-shop scheduling.
\item Local Search (LS). It is the basic local search with the same operators used in EoH. 
\item ILS1~\cite{stutzle1998applying}: It is an iterated local search developed for FSSP.
\item ILS2: It uses the same framework as ours but with hand-crafted heuristic strategy.
\end{itemize}


We evaluate the algorithms on the widely-used Taillard Instances. We test 11 different test sets.  The number of jobs in these instances ranges from 20 to 200, and the number of machines ranges from 5 to 20.

Table~\ref{table:taillard_appendix} presents the results obtained from different algorithms on Taillard instances. The table gives the average gaps to the upper bounds provided in~\citet{taillard1993benchmarks}. The average is calculated on 10 instances for each test set. The best results are highlighted in bold. \ourmethod{} is the best on most test sets and obtain the best average gap of 0.23\%. \ourmethod{} outperforms these commonly used heuristics as well as the recent deep learning neural solvers. Notably, \ourmethod{} outperforms ILS2, which shares the same framework and local search operators but uses a human hand-crafted heuristic strategy for perturbation.

\begin{table}[t]
\centering
\caption{Results on Taillard instance sets. The value is the average gap to the best-known solutions on 10 instances in each set. The best results are in bold. }~\label{table:taillard_appendix}

\resizebox{0.7\textwidth}{!}{%
\begin{tabular}{lccccccccc}
\toprule

  Test Set& GUPTA & CDS   & NEH  & NEHFF & PFSPNet & LS   & ILS1 & ILS2  & \ourmethod{}    \\
\midrule
20\_5   & 12.89 & 9.03  & 3.24 & 2.30  & 2.30 & 1.91 & 0.42 & 0.18  & \textbf{0.09}  \\
20\_10  & 23.42 & 12.87 & 4.05 & 4.15  & 4.04 & 2.77 & 0.33 & \textbf{0.25} & \textbf{0.30}  \\
20\_20  & 21.79 & 10.35 & 3.06 & 2.72  & 2.96 & 2.60 & 0.29 & 0.25  & \textbf{0.10}  \\
\midrule
50\_5   & 12.23 & 6.98  & 0.57 & 0.40  & 0.51 & 0.32 & 0.15 & 0.32  & \textbf{0.02}  \\
50\_10  & 20.11 & 12.72 & 3.47 & 3.62  & 3.48 & 3.33 & 1.47 & 0.29  & \textbf{0.19}  \\
50\_20  & 22.78 & 15.03 & 5.48 & 5.10  & 5.05 & 4.67 & 2.13 & \textbf{0.34} & 0.60   \\
\midrule
100\_5  & 5.98  & 5.10  & 0.39 & 0.31  & 0.31 & 0.28 & 0.20 & 0.38  & \textbf{-0.04} \\
100\_10 & 15.03 & 9.36  & 2.07 & 1.88  & 1.72 & 1.38 & 0.77 & 0.34  & \textbf{0.14}  \\
100\_20 & 21.00 & 13.55 & 3.58 & 3.73  & 3.56 & 3.51 & 2.27 & \textbf{0.43} & \textbf{0.41}  \\
200\_10 & 11.59 & 7.22  & 0.98 & 0.70  & 0.82 & 0.87 & 0.74 & 0.54  & \textbf{0.12}  \\
\midrule
200\_20 & 18.09 & 11.89 & 2.90 & 2.52  & 2.49 & 2.53 & 2.26 & \textbf{0.59} & \textbf{0.61}  \\
\midrule
Average & 16.81 & 10.37 & 2.71 & 2.49  & 2.48 & 2.20 & 1.00 & 0.36  & \textbf{0.23} \\
\bottomrule
\end{tabular}%
}
\end{table}

\end{document}